\documentclass[11pt]{article}

\usepackage[preprint]{acl}

\usepackage{textcomp}  
\usepackage{scalerel}  
\def\uhel{\textsuperscript{\scalerel*{\includegraphics{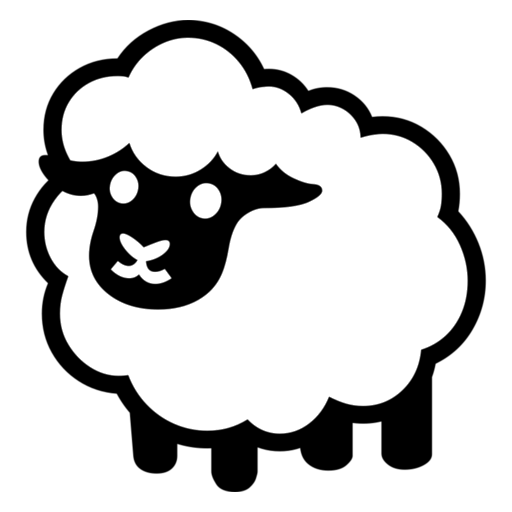}}{\textrm{\large\textbigcircle}}}}
\def\ptor{\textsuperscript{\scalerel*{\includegraphics{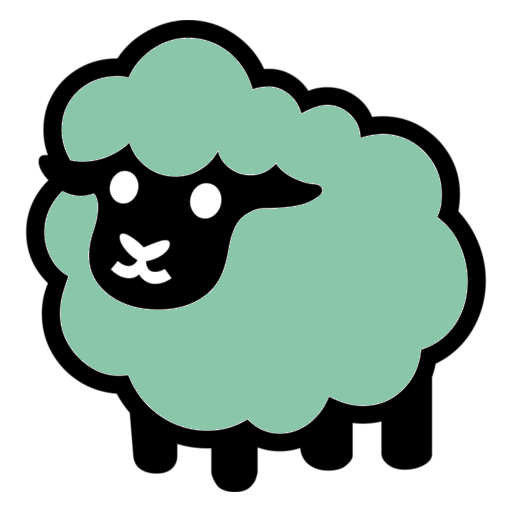}}{\textrm{\large\textbigcircle}}}}
\def\ubs{\textsuperscript{\scalerel*{\includegraphics{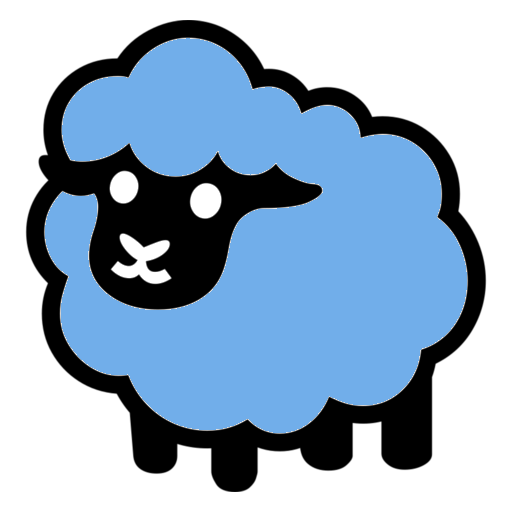}}{\textrm{\large\textbigcircle}}}}
\def\uutr{\textsuperscript{\scalerel*{\includegraphics{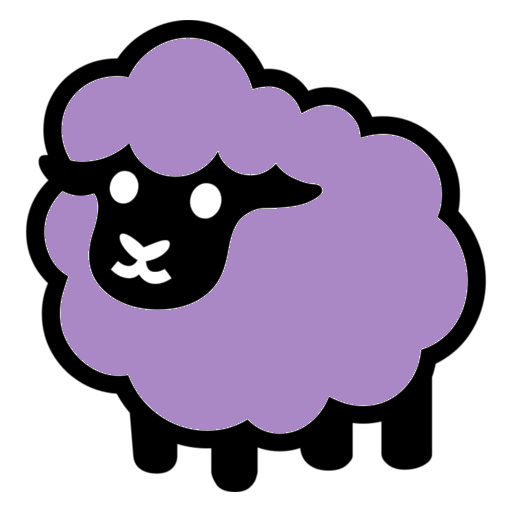}}{\textrm{\large\textbigcircle}}}}
\def\ucop{\textsuperscript{\scalerel*{\includegraphics{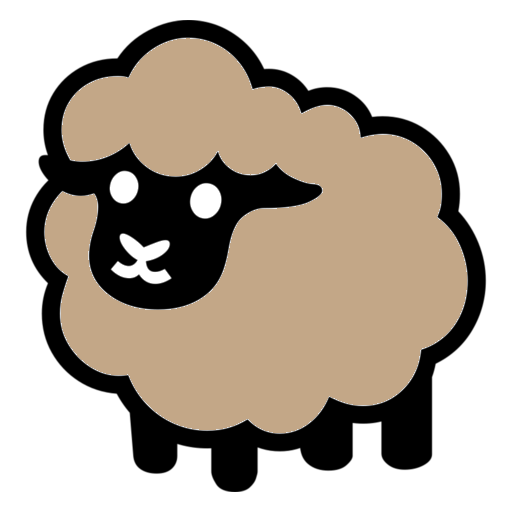}}{\textrm{\large\textbigcircle}}}}
\def\ugre{\textsuperscript{\scalerel*{\includegraphics{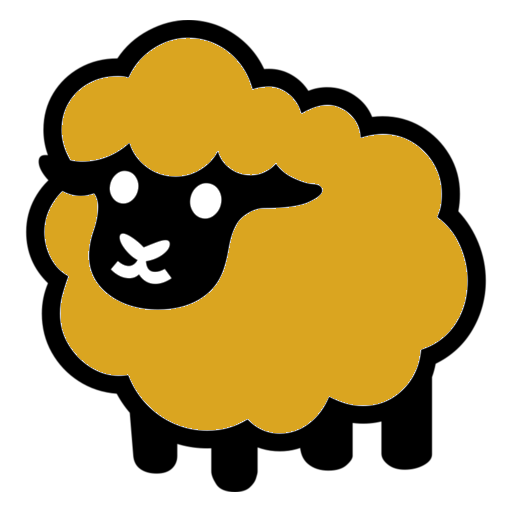}}{\textrm{\large\textbigcircle}}}}
\def\ulor{\textsuperscript{\scalerel*{\includegraphics{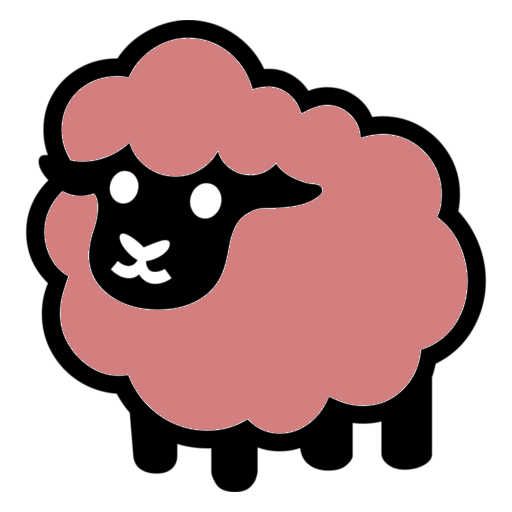}}{\textrm{\large\textbigcircle}}}}
\usepackage{times}
\usepackage{latexsym}
\usepackage{booktabs}
\usepackage{subcaption}

\usepackage[T1]{fontenc}
\usepackage{tabularx}
\usepackage{comment}
\usepackage[utf8]{inputenc}

\usepackage{microtype}

\usepackage{inconsolata}

\usepackage{graphicx}
\usepackage{enumitem}
\usepackage{relsize}
\usepackage[most]{tcolorbox}
\usepackage{cleveref}
\usepackage{todonotes}
\usepackage{subcaption}
\usepackage[export]{adjustbox}
\usepackage{siunitx}
\usepackage{multirow}
\usepackage{placeins}
\usepackage[all]{nowidow}

\setitemize{noitemsep,topsep=0pt,parsep=0pt,partopsep=0pt}

\newenvironment{instructionbox}[1]{\begin{tcolorbox}[colback=black!15,colframe=black] \small \tt \smaller #1}{\end{tcolorbox}}

\title{Can Humans Dream of Electric Sheep? Human-Written Samples for Fine-Grained Vision-and-Language Hallucination Benchmarking}

\author{Timothee Mickus\uhel \hfill Claudio Savelli\ptor \hfill Eduardo Calò\uutr \hfill Emilio Raimond\ubs \hfill \textbf{Stella Frank}\ucop \\ \textbf{Hengyu Luo}\uhel \hfill \textbf{Flavio Giobergia}\ptor \hfill \textbf{Vincent Segonne}\ubs \hfill
\textbf{Chuyuan Li}\ugre \hfill \textbf{Aman Sinha}\ulor \\
\textbf{Lorenzo Vaiani}\ptor \qquad \textbf{Jörg Tiedemann}\uhel \qquad \textbf{Raúl Vázquez}\uhel
\\[0.2cm] 
\uhel University of Helsinki \hfill \ptor Politecnico di Torino \hfill \uutr Universiteit Utrecht \hfill
\ubs Université Bretagne Sud \\\ \ucop University of Copenhagen \hfill \ugre University Grenoble Alpes \hfill \ulor University of Lorraine
}

\begin{document}
\maketitle
\begin{abstract} 
In an age of rapid model turnover, 
how do we make hallucination evaluation more perennial?
We explore whether human-written hallucination samples could take the place of model-generated hallucinations, in order to make benchmarking detection independent of particular models.
To this end, we construct SHEEP, a dataset of
1,600 human-written samples, spanning four languages (Chinese, English, French, Italian), and 18,400 samples from five vision-and-language models, all annotated for hallucinations using a fine-grained span-level labeling scheme.
We find that human-written samples result in higher agreement and allow greater control of dataset contents, while remaining distributionally similar to samples derived from vision-and-language samples and providing a reasonable portrayal of detection capabilities --- suggesting that human data is a viable substitute for model-based hallucination benchmarks. 
\end{abstract}

\section{Introduction}

Reliably detecting LLM hallucinations --- outputs that seem plausible, look fluent, but are ultimately factually incorrect \cite[e.g.,][]{rawte-etal-2023-troubling, huang_survey_2025} --- is an urgent problem for NLP.
This is even more problematic for visually-grounded language models: they have specific hallucination propensities related to image misunderstanding, along with the pathologies inherited from their language backbones~\citep{liu2024surveyhallucinationlargevisionlanguage}.


A key part of the solution is developing accurate and robust benchmarks and evaluation practices \cite[e.g.,][]{vazquez-etal-2025-semeval, niu-etal-2024-ragtruth}.
Here, a fundamental challenge is the difficulty of developing hallucination detection benchmarks that are independent of the specific models used to generate the benchmark data.
Indeed, most hallucination benchmarks are deeply dependent on specific models, either as sources of faulty outputs to be annotated \citep[e.g.,][]{ravichander-etal-2025-halogen, gamba-etal-2026-confabulations}, or as tools to produce synthetic data that emulates the phenomenon \citep[e.g.,][]{muhlgay-etal-2024-generating}. As a result, these benchmarks make it difficult to discern whether good detection performance is due to correct understanding of hallucinations or simply to overfitting to idiosyncrasies of a specific model.
The problem is compounded by the rapid pace of development of modern NLP: a dataset compiled using a popular model may fail to represent contemporary systems by the time of publication, rendering its findings potentially obsolete \citep[e.g.,][]{laskar-etal-2023-systematic, Liu2025-df}. 

Beyond this first set of issues, the construction of hallucination benchmarks itself is rife with practical hurdles. Randomly sampling outputs from a foundation model is representative of model behavior, but results in sparse representation of rare hallucination types, which may cause problems for evaluating a detection method's ability to identify these hallucinations. More targeted sampling using a LLM-judge for preselecting items, while it may allow us to balance hallucination types somewhat, ultimately depends on the judge's accuracy, making 
the effectiveness of such setups hard to predict.

In this work, we ask whether human-generated hallucinations could be a solution to these two sets of problems. 
Human-written samples could not only allow us to precisely control the prevalence of hallucinations of any particular type, but also bypass the reliance on specific models and shelter benchmarks from obsolescence. 

We test this hypothesis in the context of large vision-and-language models (LVLMs), and propose the SHEEP dataset: a Set for Human-written and Electronic Erroneous Productions.
We define a set of hallucination types that are caused by mistakes in visual grounding, such as miscounting objects or misreading text in the image, and compare hallucination rates and hallucination type distributions in randomly sampled data, items preselected by a LVLM-judge, as well as hallucinations manually invented by humans. SHEEP spans a total of 20,000 samples equally balanced over four languages (Chinese, English, French, Italian) labeled at a span-level by three annotators.
 
Through careful analysis of the collected data, we find that human-written samples yield higher inter-annotator agreement and more controllable dataset contents, while maintaining a distributional profile similar to LLM-derived samples and providing a reasonable assessment of the  performances of detection tools on LVLM data. These observations validate 
that human-written items are a viable substitute for LVLM-based hallucination benchmarks.
Our contributions are:
\begin{enumerate}[nosep]
\item SHEEP, a multilingual dataset containing 20,000 model and human outputs; 
\item Fine-grained, multi-annotator annotations at the character-level, differentiating between five possible hallucination labels; 
\item Analyses demonstrating the viability of human-written hallucinations for hallucination-detection evaluation.
\end{enumerate}
SHEEP is available under a CC-BY-NC license at \href{https://helsinki-nlp.github.io/shroom/2026}{\tt helsinki-nlp.github.io/shroom/2026}.

\section{Related works}
\label{sec:rel-work}
\paragraph{Hallucinations in LVLMs.} Hallucinations in vision-and-language models are generally defined as content that is not supported or contradicts the visual input \citep{liu2024surveyhallucinationlargevisionlanguage, bai2025hallucinationmultimodallargelanguage}. Early works organized the phenomenon on a coarse triad of object, attribute, and relation errors \citep{liu2024surveyhallucinationlargevisionlanguage}. Later proposals refine this scheme in multiple directions: \citet{jiang2024haleval} add event hallucinations to account for entirely fabricated scenes, while \citet{rani2024visualhallucinationdefinitionquantification} distinguish up to eight fine-grained classes, including miscounting, misreading of visible text, and identity incongruity.
Our own 
classification (\Cref{sec:framework}) draws from this line of work, but is deliberately kept compact so that it can be applied reliably at the span level. 

\paragraph{LVLM hallucination benchmarks.} Existing benchmarks for LVLM hallucination broadly fall into three paradigms \citep{liu2024surveyhallucinationlargevisionlanguage}. 

\emph{1. Caption-centric} 
approaches assess the factual accuracy of model-generated descriptions or answers. 
Metrics such as CHAIR \citep{rohrbach2018object} and its open-vocabulary extension \citep{petryk-etal-2024-aloha} remain restricted to object-level errors. 

\emph{2. Discriminative} approaches reframe evaluation as a classification task to detect preexisting hallucinations in image-text pairs. 
\citeposs{shekhar-etal-2017-foil} seminal work inserted single-word substitutions into the gold captions to produce a benchmark.
\citet{li-etal-2023-evaluating} established the standard for object hallucination detection via yes/no probing about object presence. Recent variants have improved 
reliability by addressing dataset biases and devising novel 
consensus strategies \citep{lovenia-etal-2024-negative, pham2024hpopehierarchicalpollingbasedprobing}.

\emph{3. Hybrid} approaches integrate both generative and verification tasks. As such, this category concentrates on works that attempt to holistically measure model reliability from different standpoints. Some works target entangled visual illusions and language hallucinations \citep{Guan_2024_CVPR}, others focus on perturbing input data for hallucination detection 
\citep{ding2024hallupi,saito2026haldecbench}, and yet others automate benchmark generation \citep{wu2024autohallusion}. Another line of work 
focuses on fine-grained classifications of failures, e.g., \citet{ye2024beaf} introduce image edits for true understanding, ignorance, stubbornness, and indecision. \citet{wang2024amberllmfreemultidimensionalbenchmark} offer an LLM-agnostic evaluation pipeline constructing a dataset where answers are easily verifiable with perhaps the closest taxonomy to the one we propose. Also closely related to our setup are M-HalDetect \citep{gunjal2024mhaldetect}, providing 16k fine-grained annotations over model responses
, and HalLoc \citep{Park_2025_CVPR} offering token-level hallucination localization. 

\paragraph{Open problems.} The vast majority of these resources are built around English outputs from one or a handful of LVLMs. This raises two concerns that directly motivate our work. First, benchmarks based on LVLM generations inherit the characteristics of their source models, and their diagnostic value degrades as the underlying systems are replaced \citep{laskar-etal-2023-systematic, Liu2025-df}, a recurring theme also observed in LLM-centric resources \citep{ravichander-etal-2025-halogen, vazquez-etal-2025-semeval}. Efforts to sidestep this via synthetic generation \citep{muhlgay-etal-2024-generating} still rely on an LLM as a generator.
Our work addresses both gaps, with span-level annotations over five recent LVLMs and a human-written subset across four languages.

\begin{table*}[th]
\centering
\small
\begin{tabularx}{\linewidth}{c@{{~~}}lX}
\toprule
\multicolumn{2}{l}{\textbf{Label}} & \textbf{Description} \\
\midrule
\textbf{A}&{Invention} & An entity, object, property, or event that is not present in the image. \\
\textbf{B}&{Mischaracte\-ri\-za\-tion} & An entity, object, property, or event present in the image but described incorrectly. \\
\textbf{C}&{OCR Problem} & Misreading text that is visible in the image. \\
\textbf{D}&{Miscounting} & Incorrectly reporting the quantity of visible items. \\
\textbf{E}&{Other} & Does not correspond to any of the previous classes (use sparingly). \\
\midrule
\textbf{F} & None & (\textit{Unused by annotators when marking spans; inferred from the lack of annotation for an item.}) \\
\bottomrule
\end{tabularx}
\caption{Hallucination classification labels.}
\label{tab:hallucination_labels}
\end{table*}

\section{Theoretical framework}
\label{sec:framework}

We focus on devising a classification of hallucinations that can serve as a theoretical basis for dataset creation and that annotators can reliably follow during annotation. The main issues we consider in designing this classification are (1) \textit{error coverage} and (2) \textit{category granularity}. We aim to be as comprehensive as possible by covering a wide range of errors, while avoiding overly fine-grained distinctions that would introduce unnecessary confusion. 

We define hallucination as content that is either unsupported by or contradictory to the semantic reference, which is the input image in our case. 
Therefore, we focus on output correctness rather than coverage: is the model generation true with regard to the image, not whether everything in the image has been described.
We only consider single-turn interactions, 
focusing on open-ended tasks such as visual question answering or image captioning. 

We develop the classification scheme outlined in \Cref{tab:hallucination_labels}, drawing inspiration from previous benchmarks. 
We reorganize previously proposed taxonomies, merging categories that we consider similar (e.g.,~\textit{contextual guessing} in \citealp{rani2024visualhallucinationdefinitionquantification} and \textit{existence} in \citealp{yan2026measuringmeasurersqualityevaluation}), and removing categories that are not relevant to our purposes (e.g.,~\textit{gender anomaly} in \citeauthor{rani2024visualhallucinationdefinitionquantification}). 

We encountered several challenges during the development of our classification, particularly regarding grounding criteria (i.e.,~whether outputs should rely solely on visual evidence or may also incorporate external knowledge), and interpretative subjectivity (i.e.,~cases where the output may or may not be inferable depending on the annotator's background). To address these ambiguities, 
rather than creating separate categories, we instruct annotators to adhere to our definition of hallucination. Thus, factual claims unverifiable from the image are treated as a hallucination; 
expressions of uncertainty (e.g.,~``it looks like'') are not to be marked as hallucinations as this is a desirable 
behavior. 

\section{Data collection}
\label{sec:data}




We generate outputs using two existing datasets.
HaloQuest~\citep{10.1007/978-3-031-72980-5_17} is a 
VQA dataset designed to provoke hallucinations, for example with visually challenging images or false-premise questions not matching the image. We use only the non-synthetic images.
VISaGE~\citep{frank-allaway-2025-visage} contains non-synthetic images of objects with unusual characteristics (e.g.,~three-legged cats, boat ambulances), along with questions about these characteristics.
%
%
To obtain LVLM outputs which may include hallucinations, we use five models:
Gemma3 \citep{gemmateam2025gemma3technicalreport}, 
InternVL3 \citep{chen2024internvl}, 
MiniCPM V 4.5 \citep{yu2025minicpmv45cookingefficient}, 
Llava-NeXT \citep{liu2024llavanext}, 
and Qwen3-VL \citep{Qwen3-VL}. 
All models 
have 8B parameters except Gemma3 with 27B.
We use a temperature of $\tau=0.7$ and a maximum 
length of 512 tokens, across 5 random seeds. 
To encourage Gemma3 to generate responses in the same language as the input query, 
we localize the system prompt. 
We translate the two source datasets 
using NLLB-200 3.3B \citep{nllbteam2022languageleftbehindscaling} for Italian and French, and using Qwen3~8B \citep{qwen3technicalreport} for Chinese; 
see \Cref{adx:translation} for details. 

\paragraph{Sampling strategies.}
We select the data for human annotation using three sampling strategies:
(1)~random, (2)~model-assisted pre-selection, and (3)~human-written.
For the first two, we select items from the LVLM-generated outputs. For the third, we collect human-invented responses that include hallucinations to create a model-independent benchmark for hallucination detection.

\textit{1.~Random.}
From the generations for HaloQuest and VISaGE, we randomly select 460 samples for each language and model pair, for 2,300 samples per language in total.

\textit{2.~Model-assisted pre-selection (MAP).} 
As a comparison point for our intended human-created benchmark, we test whether an LVLM-guided sampling process can lead to a high quality evaluation benchmark.
We use Gemma3~27B~IT to pre-select LVLM-generated outputs for annotation. We run Gemma3 on all our outputs to assess which outputs contain hallucinations, using our classification in \Cref{tab:hallucination_labels}. For each datapoint, the judge receives the input image, the question, and the complete LVLM response, together with a prompt containing a role description, label definitions, and formatting instructions.\footnote{See \Cref{fig:prompt-llm-judge} in \Cref{adx:guidelines} for prompt instructions.} The instructions closely match the human guidelines we provide to annotators. 
We then use these preliminary labels to construct a more label-balanced annotation set.
In practice, the judge predicts OCR problems and miscounting much less frequently than invention and mischaracterization.
We therefore include all datapoints predicted as OCR problems or miscounting, and sample from the remaining predicted labels so as to balance invention and mischaracterization as much as possible. This yields 2,300 MAP samples per language. 

\textit{3.~Human-written samples.} 
Finally, we collect human-written samples as a model-independent source of hallucination examples.
For each language, we (i)~randomly select 100 image-question inputs from the data we used to generate LVLM outputs; (ii)~have writers fluent in the language manually create outputs that deliberately contain mistakes akin to hallucinations; (iii)~machine-translate these 100 samples into the other three languages;\footnote{Using the pipeline described in \Cref{adx:translation}.} (iv)~manually check translated samples and post-edit them when relevant. 
This results in a total of 1,600 human-written samples, out of which 400 are directly written by humans and 1,200 are machine-translated and manually curated. This manual curation step proved necessary as we observed that the NLLB backbone used for FR and IT had a tendency to ignore anything but the first sentence. We provide a brief overview of the improvement in quality, as measured by COMETKiwi scores (\citealp{rei-etal-2022-cometkiwi}\footnote{\texttt{Unbabel/wmt22-cometkiwi-da} model} in \Cref{tab:trans-postedit-quant}: while absolute values are hard to interpret, we confirm that post-editing systematically improves quality estimation scores. 

\begin{table}
    \centering
    \begin{subtable}{\columnwidth}
\centering
\small\smaller
    \sisetup{
      table-format=1.3,
      round-mode=places,
      round-precision=3
    }
    \begin{tabular}{>{\bf}l *{4}{S}}
    \toprule
         &  {{\textbf{to EN}}}&  {{\textbf{to FR}}}&  {{\textbf{to IT}}}&  {{\textbf{to ZH}}}\\
    \midrule
        from EN & {{--}} & 0.7991 &	0.8174 	& 0.7797 \\
        from FR & 0.8442 & {{--}} & 0.808 &	0.7871\\
        from IT & 0.8204 & 0.8153 &	{{--}} &	0.815 \\
        from ZH & 0.8282 & 0.7715 &	0.8029 	& {{--}} \\
    \bottomrule
    \end{tabular}
    \caption{Before post-editing}
    \end{subtable}
    \begin{subtable}{\columnwidth}
\centering
\small\smaller
    \sisetup{
      table-format=1.3,
      round-mode=places,
      round-precision=3
    }
    \begin{tabular}{>{\bf}l *{4}{S}}
    \toprule
         &  {{\textbf{to EN}}}&  {{\textbf{to FR}}}&  {{\textbf{to IT}}}&  {{\textbf{to ZH}}}\\
    \midrule
        from EN & {{--}} & 0.8492 & 0.8511 & 0.8289\\
        from FR & 0.8459 &{{--}} & 0.8111 & 0.7981\\
        from IT & 0.8485 & 0.8553 &{{--}} & 0.8177\\
        from ZH & 0.8306 &	0.7881 &	0.8076 &{{--}} \\
    \bottomrule
    \end{tabular}
    \caption{After post-editing}
    \end{subtable}
    \caption{Quality improvement from post-editing}
    \label{tab:trans-postedit-quant}
\end{table}

Writers were familiar with the annotation task: they participated in pilot experiments and had access to the same 10 throwaway training items we provide to annotators.
They were asked to generate a single type of hallucination for each sample (see~\Cref{sec:framework}), and label each sample with the hallucination type. The collection interface tracked the distribution of intended hallucination types to encourage an even distribution across labels. 

\paragraph{Annotation.}
We recruited 12 C1-to-native proficient annotators per language, mostly European university students.\footnote{Due to one dropout, French has 11 annotators.} To obtain three judgments per sample, each annotator processed $\sim$1,250 items over one month, compensated at $\sim$€15/h.\footnote{29 items have 4 annotations due to assignment collisions.} The workflow included autonomous training on throwaway samples, a briefing session, and access to support via a collaborative FAQ and direct means of contact.
Annotators could self-pace and edit previous entries. 
See annotation and writing guidelines and interface details in \Cref{adx:guidelines}. 

\begin{figure}[!t]
    \centering
\begin{subfigure}{\linewidth}
\centering
\includegraphics[max width=0.7\linewidth]{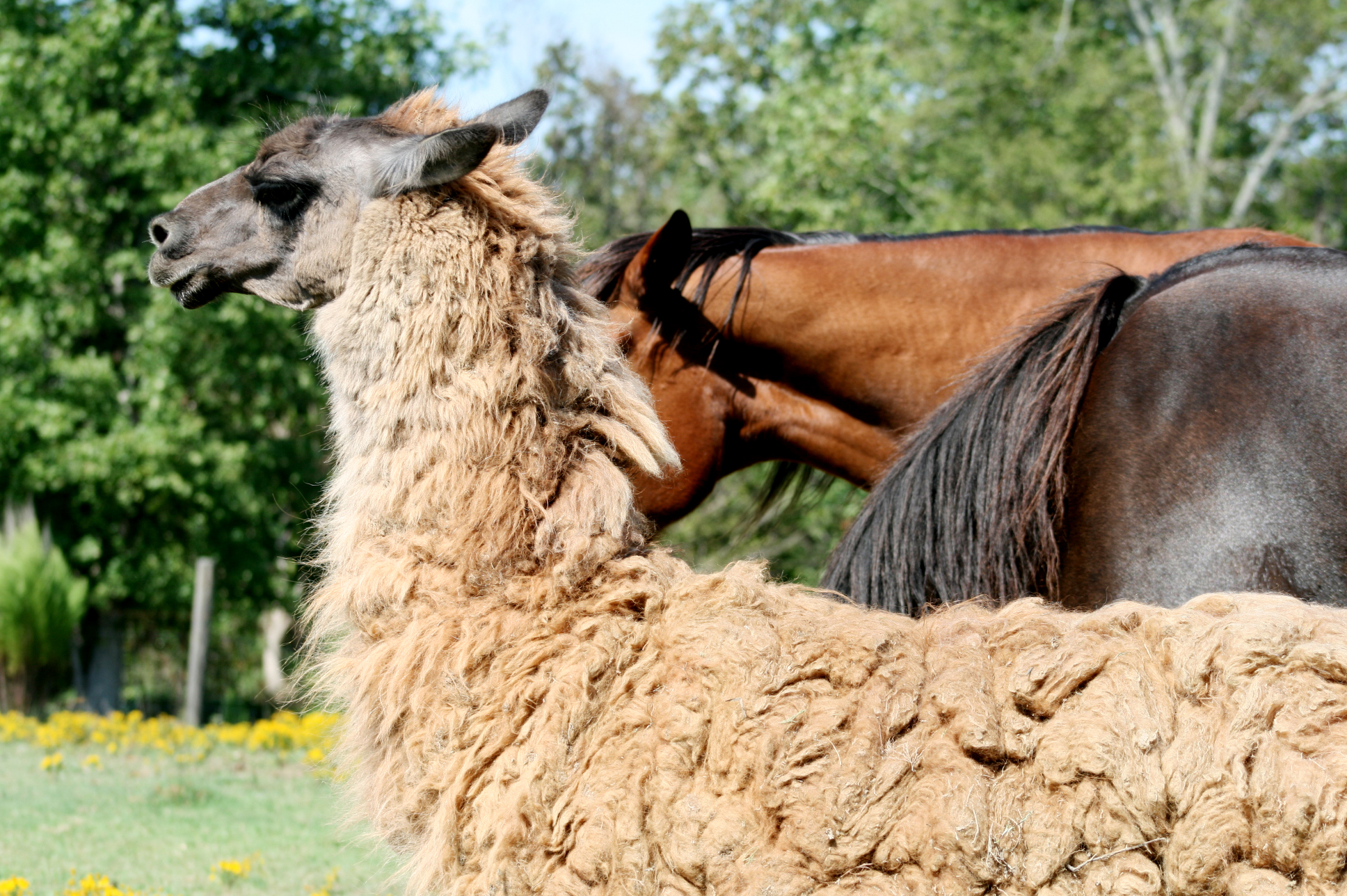}
\caption{Associated image from HaloQuest \citep{10.1007/978-3-031-72980-5_17}.}
\end{subfigure}
\begin{subfigure}{\linewidth}
\input{latex/figs/example/train-fr-4186}
\vspace{-0.75em}
\subcaption{Annotated datapoint; the model is prompted to describe what the black sheep is doing, the response claims it is grazing.}
\end{subfigure}
    \caption{Example SHEEP datapoint.}
    \label{fig:datapoint}
\end{figure}

\paragraph{Example datapoint.} We include an example datapoint in \Cref{fig:datapoint} for reference. We provide both individual and aggregated annotations, along with comments and relevant metadata. 

\section{Analyses}
\label{sec:results}

We assess whether our human-written samples are a viable substitution for an LVLM-generated benchmark in two ways: we look at characteristics of the collected data (inter-annotator agreement, ability to surface specific hallucination types, distributional gaps) in \Cref{sec:results:qc}, and then compare evaluations of hallucination detectors in \Cref{sec:results:benchmark}.

\subsection{Data characteristics}
\label{sec:results:qc}

\begin{figure}
    \centering
    \includegraphics[max width=0.9\linewidth, trim={0 0.5cm 0 0.5cm}, clip]{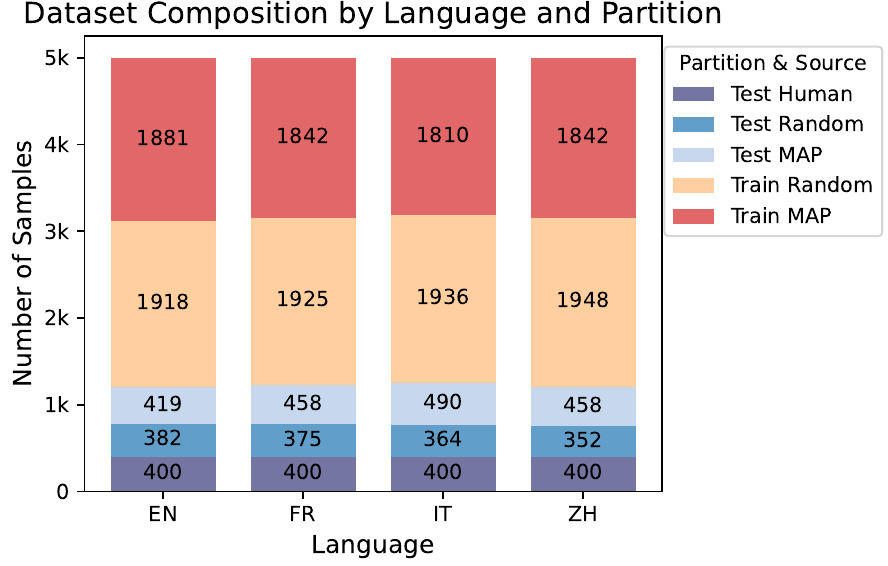}\\ 
    \includegraphics[max width=0.85\linewidth]{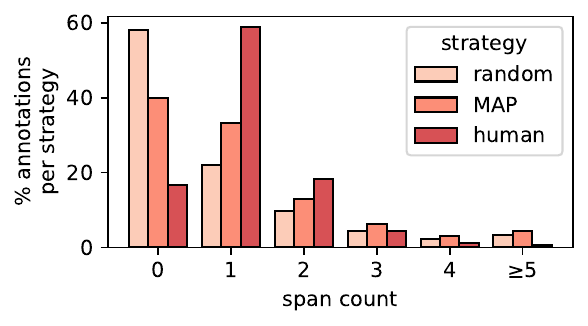}
    \caption{\textit{Top:} Samples per strategy, language, and split. \textit{Bottom:} Proportion of annotations grouped by number of hallucination spans marked in an item.}
    \label{fig:counts}
\end{figure}

\paragraph{Descriptive statistics.}
In \Cref{fig:counts}~(top), we show the number of samples for each language grouped by sampling strategy.\footnote{Images used in human-written data are all in the test split.} \Cref{fig:counts}~(bottom) presents the distribution of the number of hallucination spans annotated per item, again by sampling strategy. Annotators rarely mark more than one hallucination span per sample: the distributions for the random and MAP strategies are approximately Zipfian, with most annotations marking no spans, while annotations with five or more spans are  rare ($\sim$3.5\%). In contrast, the human-written samples contain a larger amount of annotations with exactly one span, whereas the MAP data comprises more annotations with three or more spans.
This lines up with our expectations: whereas an uninformed random sample is very ineffective at surfacing hallucinations, human writers are quite capable of matching our classification; the success of the MAP is contingent on the accuracy of the LVLM-judge.

\paragraph{Inter-annotator agreement.}
We consider three metrics to assess inter-annotator agreement (IAA). 

\begin{figure*}[ht]
    \centering
    \includegraphics[width=0.9\textwidth, trim={0 0 0 0}, clip]{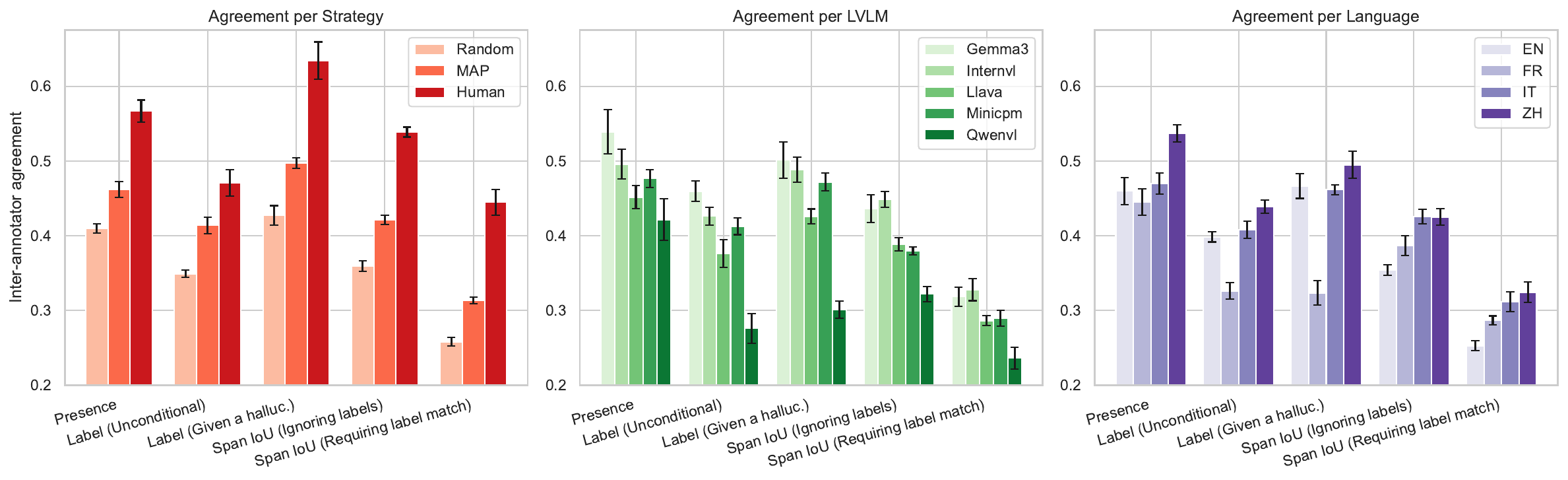} 
    \caption{Annotator agreement 
    grouped by sampling process (left), generating LVLM (center), and output language (right). Center and right plots show only LVLM-generated data; left also includes human-generated data. Error bars represent 95\% bootstrap confidence, sampling with replacement. See \Cref{adx:sup res:full IAA} for table versions. } 
    \label{fig:agreement-plot}
\end{figure*}

\textit{1. Presence agreement.} We measure whether annotators agree on which samples contain hallucinations, regardless of hallucination type and specific span of hallucinated text. 
We adopt a free-marginal version of Krippendorff's alpha: 
    $$\alpha_\mathrm{free} = 1 - (k/n)\sum{}_{v, v'}^Vo_{vv'}\delta(v, v') $$
where $n$ is the total number of paired ratings, $k$ is the number of labels (i.e., 2), $\delta$ is the nominal metric, and $o$ is the matrix of observed coincidences. We use the free-marginal formulation because it is less sensitive to differences in label distributions arising from varying rates of hallucinations \citep{Byrt1993-ra,randolph-2005-free}. This property is particularly important as the three sampling strategies are designed to yield different hallucination rates.
    
\textit{2.~Label agreement. } We quantify to what extent annotators agree on the hallucination \textit{type} present in a sample, using a standard (fixed-marginal) Krippendorff's alpha. We contrast agreement across the full dataset (Unconditional) vs. what we observe only on data where all annotators agree on the presence of a hallucination (Given a hallucination).\footnote{
To handle annotations with  more than one span, we ignore all spans beyond the first. See \Cref{adx:sup res:first-vs-longest}  for discussion.}

\textit{3. Span agreement. } The overlap between spans of text marked as containing hallucinations. We use an Intersection over Union (IoU), as per prior work \citep{vazquez-etal-2025-semeval}. 
We compute the average pairwise IoU across all possible 
pairs of annotations of the same item. We compute span agreement when factoring in hallucination type in the intersection (Requiring label match) and regardless of assigned label (Ignoring labels).

IAA scores per sampling strategy are reported in \Cref{fig:agreement-plot}~(left). 
Unsurprisingly, controlling for similarity on other annotation levels (presence or absence of hallucination, choice of specific labels) improves agreement rates.
More interestingly, are the clear distinctions across strategies: annotators are most consistent on the human-written samples and least consistent on the randomly sampled datapoints, suggesting that  
sampling strategies very clearly influence the ability of annotators to reach a consensus, and that agreement improves as the selection becomes more controlled. 






A key question to assess whether human-written data can serve as a proxy for model-dependent hallucination benchmarks is how IAA varies across generating LVLMs and languages. 
Such a substitution is only meaningful if the resulting annotations exhibit levels of agreement comparable to the variation naturally induced by changes in model or language. 
To this end, \Cref{fig:agreement-plot} reports IAA on the random and MAP subsets across LVLM (center) and language (right). 
\begin{figure*}[t]
    \centering
    \begin{subfigure}[t]{0.3\linewidth}
        \centering
        \includegraphics[width=\linewidth]{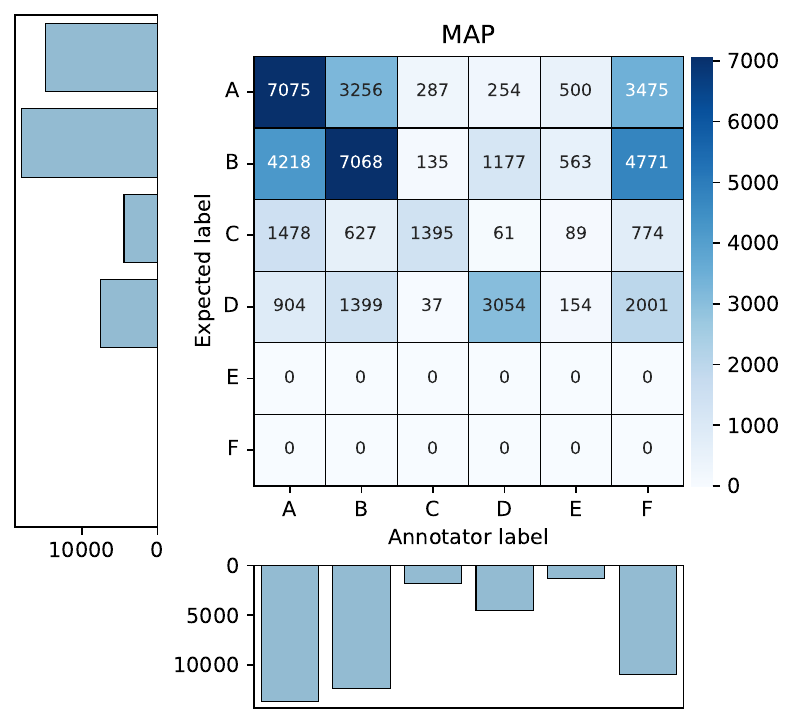}
        \caption{MAP judge labels vs.~annotator labels, \textit{MAP} data.}
    \end{subfigure}
    \hfill
    \begin{subfigure}[t]{0.3\linewidth}
        \centering
        \includegraphics[width=\linewidth]{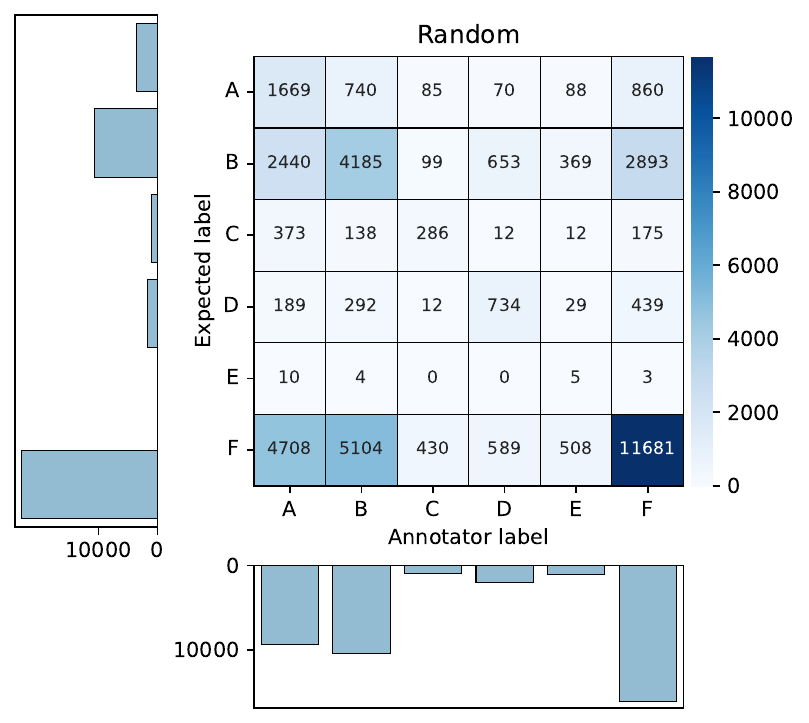}
        \caption{MAP judge labels vs.~annotator labels, \textit{random} data.}
    \end{subfigure}
    \hfill
    \begin{subfigure}[t]{0.3\linewidth}
        \centering
        \includegraphics[width=\linewidth]{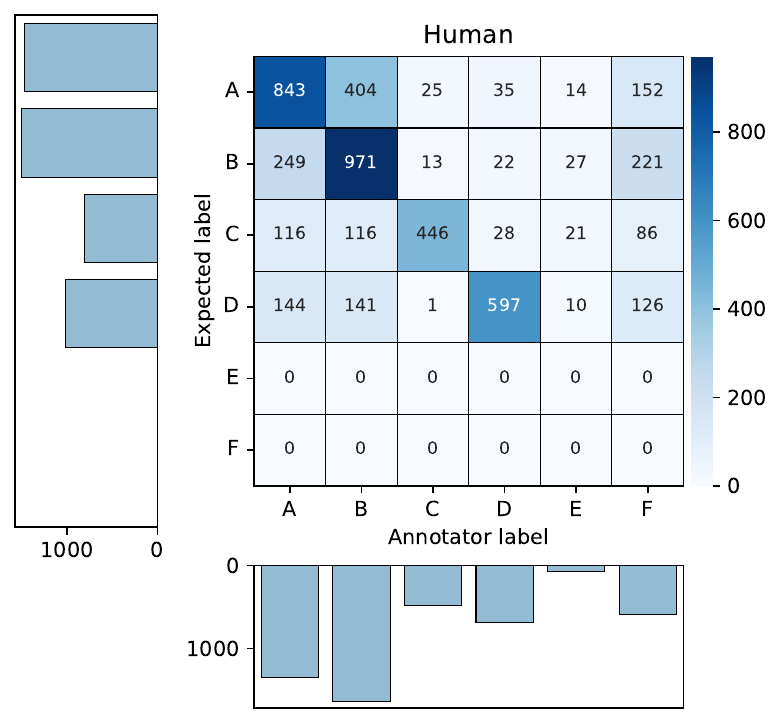}
        \caption{Writer labels vs.~annotator labels, \textit{human} data.}
    \end{subfigure}
    \caption{Distribution of expected labels (row) vs. annotator labels (columns) for all available annotations. 
    }
    \label{fig:external_labels}
\end{figure*}
%
None of the generating LVLMs in \Cref{fig:agreement-plot}~(center) yields agreement levels higher than those observed for the human-written samples in \Cref{fig:agreement-plot}~(left); 
and variation across generating models is often as large as, or larger than, the variation induced by the sampling strategies themselves. Among the evaluated models, annotators tend to agree most on samples generated by the largest model (Gemma3), whereas QwenVL consistently produces lower agreement scores. Similarly, across the four languages in \Cref{fig:agreement-plot} (right), we observe a similar pattern: agreement varies substantially, with French consistently exhibiting lower IAA and Chinese yielding the highest.

Overall, these results suggest that the increase in agreement observed for human-written samples falls within the range of variation induced by changes in the generating LVLM or the language.

\paragraph{Agreement with expected labels.} 
Do annotators agree with the MAP-judge assigned hallucination labels or the human writers of hallucinations?
%
%
%
\begin{table}[tb]
\centering
\small\smaller

\begin{tabular*}{0.9\columnwidth}{l@{\extracolsep{\fill}}cccc}
\multicolumn{5}{c}{\textbf{3-way classification}} \\
\hline
\textbf{Origin} & \textbf{EN} & \textbf{FR} & \textbf{IT} & \textbf{ZH} \\
\hline
\textbf{LVLM-visage}    & $\mathbf{0.958}$ & $\mathbf{0.862}$ & \underline{$0.859$} & $\mathbf{0.857}$ \\
\textbf{LVLM-haloq.} & \underline{$0.878$} & \underline{$0.847$} & $\mathbf{0.869}$ & $0.757$ \\
\textbf{human}           & $0.867$ & $0.743$ & $0.738$ & \underline{$0.770$} \\
\hline
\end{tabular*}

\vspace{0.4em}

\begin{tabular*}{0.9\columnwidth}{l@{\extracolsep{\fill}}cccc}
\multicolumn{5}{c}{\textbf{6-way classification}} \\
\hline
\textbf{Origin} & \textbf{EN} & \textbf{FR} & \textbf{IT} & \textbf{ZH} \\
\hline
\textbf{gemma3}   & $0.857$ & $0.696$ & $0.595$ & \underline{$0.824$} \\
\textbf{internvl} & $0.429$ & $0.477$ & $0.495$ & $0.474$ \\
\textbf{llava}    & $0.718$ & $0.615$ & $0.654$ & $0.703$ \\
\textbf{minicpm}\hphantom{0123456}  & $0.721$ & $0.633$ & $0.655$ & $0.614$ \\
\textbf{qwenvl}   & $\mathbf{0.877}$ & $\mathbf{0.832}$ & \underline{$0.739$} & $\mathbf{0.864}$ \\
\textbf{human}    & \underline{$0.876$} & \underline{$0.748$} & $\mathbf{0.760}$ & $0.756$ \\
\hline
\end{tabular*}

\caption{Per-class F1 scores for 3-way and 6-way classification across languages.}
\label{tab:f1-combined}
\end{table} 
\Cref{fig:external_labels} quantifies the extent to which annotations agree with these expected labels.
%
The main departures from agreement along the diagonal are due to two patterns:
first, confusion between \textit{invention} and \textit{mischaracterization}, particularly in the MAP and human-written subsets,  
and secondly, human annotators disagreeing with the labels provided by the MAP and human writers (F columns).\footnote{
    Disagreement between writers and annotators is to be expected, given the IAAs reported in \Cref{fig:agreement-plot}: what counts as a hallucination is oftentimes subjective \citep{mickus-etal-2024-semeval}. This also underscores there are limits in terms of subtlety in judgment we can expect from  annotators.
}
We also observe a high false positive rate in terms of binary detection for our LVLM-judge, much higher than what we observe for human-written samples. 
In contrast, the random data contains a large proportion of correctly identified non-hallucinations; however the MAP judge labels also miss many hallucinations found by annotators (F row).
We see that the human-written sampling strategy is more effective at targeting specific hallucination types, since it avoids the high false negative and false positive rates observed when using the MAP strategy with LVLM-judge labels. 

\paragraph{Distributional shifts.}



To assess distributional shift between human-written and LVLM-generated data, we train a logistic regression classifier using character- and word-level n-gram features to predict the origin of each sample in multi-way settings, and report 5-fold cross-validation results in \Cref{tab:f1-combined}.

In the \textbf{3-way setting} (human-written, LVLM responses on VISaGE and HaloQuest), human-written samples are usually distinguishable from generated data, suggesting a clear distributional shift. However, this separability is comparable to that observed between the two LVLM benchmarks themselves (VISaGE vs. HaloQuest). In the \textbf{6-way setting}, where responses are grouped by generating LVLM, human-written data remains identifiable but is not uniquely distinct: Gemma3 and QwenVL are equally or even more easily separable based on textual features alone. 
Both results suggest that stylistic and lexical variation across LVLMs is already substantial, and that the distributional gap between human-written and LVLM-generated samples is comparable to variation across LVLMs.

\subsection{Usefulness as a benchmark}
\label{sec:results:benchmark}

Thus far, we have discussed whether human-written samples have characteristics that would differ significantly from those of LVLM-generated items. 
We have yet to assess the usefulness of human-written samples when it comes to benchmarking performances on LVLM hallucination detection. 
We evaluate a modern hallucination detection model, HalluShift++ \citep{nath2025hallushift++}, as well as our Gemma3 LLM-judge on our data.

\begin{table}[t]
    \centering
    \resizebox{\linewidth}{!}{
    \sisetup{
  table-format=1.3,
  round-mode=places,
  round-precision=3
}
\begin{tabular}{>{\bf}l@{{~}}>{\bf}l*{6}{S}}    
\toprule
 & & {\textbf{human}} & {\textbf{gemma3}} & {\textbf{internvl}} & {\textbf{llava}}  & {\textbf{minicpm}}  & {\textbf{qwenvl}} \\
\midrule
\multirow{2}{*}{\rotatebox{90}{all}}
& acc & 0.664375 & 0.542763 & 0.531915 & 0.574468 & 0.485437 & 0.459459 \\
& F1  & 0.498913 & 0.442032 & 0.400734 & 0.455147 & 0.409673 & 0.426878 \\
\midrule
\multirow{2}{*}{\rotatebox{90}{EN}}
& acc & 0.695000 & 0.521739 & 0.516129 & 0.567901 & 0.448718 & 0.402174 \\
& F1  & 0.510781 & 0.391259 & 0.328004 & 0.515805 & 0.344703 & 0.306616 \\
\midrule
\multirow{2}{*}{\rotatebox{90}{FR}}
& acc & 0.677500 & 0.539474 & 0.463415 & 0.538462 & 0.480519 & 0.426667 \\
& F1  & 0.509476 & 0.449839 & 0.321223 & 0.448990 & 0.420988 & 0.458223 \\
\midrule
\multirow{2}{*}{\rotatebox{90}{IT}}
& acc & 0.650000 & 0.539474 & 0.618421 & 0.676056 & 0.518519 & 0.433333 \\
& F1  & 0.482303 & 0.451861 & 0.383105 & 0.516484 & 0.437302 & 0.458120 \\
\midrule
\multirow{2}{*}{\rotatebox{90}{ZH}}
& acc & 0.635000 & 0.566265 & 0.532258 & 0.507692 & 0.493151 & 0.594203 \\
& F1  & 0.491000 & 0.505675 & 0.505234 & 0.390369 & 0.393317 & 0.424621 \\
\bottomrule
\end{tabular}
}
    \caption{Gemma3 classification: F1 and accuracy.}
    \label{tab:gemma}
\end{table}

\paragraph{Label assessment with Gemma3 judge.} 
Our first baseline is the Gemma3 LVLM we used earlier as part of the preselection MAP strategy.
We now use this model to label items in the random and human-written samples with the six possible labels in \Cref{tab:hallucination_labels}: note that the task here is to label items, not spans.
For evaluation, we mark the item as correct if at least one annotator provided the same label as predicted by Gemma3.
The results are shown \Cref{tab:gemma}, broken down by item language and item generation type (model or human),
with full results provided in \Cref{adx:gemma}.
Overall, we see that Gemma3 labels the human-written samples with the highest accuracy, compared to the models; somewhat surprisingly, this holds true even when Gemma3 is the source model.
However, human-written samples are not completely outside the range of model samples, especially when looking at F1, precision, or precall 
rather than accuracy.
Note that we \textit{expect} accuracy to shift if detectors are not equally capable on all hallucination types.

\begin{table}[ht]
    \centering
    \small\smaller
    \sisetup{
      table-format=1.3,
      round-mode=places,
      round-precision=3
    }
    \begin{tabular}{>{\bf}l *{5}{S}}
    \toprule
         &  {{\textbf{gemma3}}} &	{{\textbf{internvl}}} &	{{\textbf{llava}}} & {{\textbf{minicpm}}} & {{\textbf{qwenvl}}} \\
    \midrule
\textbf{human}    & 0.796& 0.770 & 0.916 & 0.867 & 0.708 \\
\textbf{gemma3}   &      & 0.717 & 0.771 & 0.700 & 0.769 \\
\textbf{internvl} &      &       & 0.658 & 0.800 & 0.660 \\
\textbf{llava}    &      &       &       & 0.831 & 0.683 \\
\textbf{minicpm}  &      &       &       &       & 0.766 \\
\bottomrule
    \end{tabular}
    \caption{Comparison of performance rankings for human- and machine written data (Pearson's $r$, $n=40$).}
    \label{tab:rankings-pearson}
\end{table}

This illustrates that the error profile is roughly conserved on the human-written data, but it still leaves open one major point: whether performance rankings from different judges would be accurately captured by our human-written data. 
To that end, we select ten LVLMs to serve as judges (Gemma 3 4b, 12b and 27b, internVL 4b, 8b and 38b, QwenVL 2b, 4b, 8b and 32b) and compare their performances on human-written vs machine-written data. 
For each judge, we compute F1 scores per language (i.e., generated by a specific LVLM or by human writers), for a total of 40 setups per data source. 
We then measure whether performances are correlated across sources, using a Pearson correlation coefficient. 
This yields the scores displayed in \Cref{tab:rankings-pearson}: crucially, we find that human-written data yields the highest correlations. 
In other words, human-written samples are as adequate as samples generated by any arbitrarily chosen LVLM. 

\paragraph{Labeled-span assessment with HalluShift++.}
Next, we consider a baseline inspired by HalluShift++, which detects hallucinations from internal LVLM signals, including representation shifts, attention patterns, and confidence-related features. We adapt this idea to our span-level localization setting by running a teacher-forced forward pass over each image--question--answer triple and extracting token-level features from an evaluator model. We extract these features from Llava-NeXT, 
yielding both a model-dependent self-probing setup 
as well as an external evaluation setup, depending on which LVLM generated the samples in our data. The features are 
used to train a token-level classifier in 
a multiclass setting where tokens are assigned to one of the labels A--F (\Cref{tab:hallucination_labels}).\footnote{More details and experiments, including with EUQ \citep{huang2026detecting}, are provided in \Cref{adx:hallushift,adx:euq}.}

\begin{table}[]
    \centering

\begin{subtable}{\linewidth}
    \centering
    \small
\sisetup{
  table-format=1.3,
  round-mode=places,
  round-precision=3
}
\begin{tabular}{>{\bf}l *{4}{S}}
\toprule
Strategy &{{\bf acc }} &{{\bf prec }} &{{\bf rec }} &{{\bf F1 }} \\
\midrule
 random  & 0.440378  & 0.225735  & 0.506065  & 0.203580 \\
 MAP  & 0.403398  & 0.240161  & 0.517390  & 0.223806 \\
human  & 0.332005  & 0.240065  & 0.442583  & 0.203390 \\
\bottomrule
\end{tabular}

\caption{HalluShift++ classification metrics across strategies. \vspace{0.125cm}}
\label{tab:hallushift:global}
\end{subtable}

\begin{subtable}{\linewidth}
    \centering
    \small
\sisetup{
  table-format=1.3,
  round-mode=places,
  round-precision=3
}
\begin{tabular}{>{\bf}l *{3}{S}}
\toprule
     &  {{\bf random}} &  {{\bf MAP}} &  {{\bf all}} \\
\midrule
  gemma3  & 0.156837 & 0.227191 & 0.188716 \\
internvl  & 0.163204 & 0.202754 & 0.186056 \\
   llava  & 0.185731 & 0.223822 & 0.212049 \\
 minicpm  & 0.182057 & 0.204835 & 0.195332 \\
  qwenvl  & 0.225453 & 0.209043 & 0.220966 \\
\bottomrule
\end{tabular}
\caption{HalluShift++ F1 per generating LVLM.}
\label{tab:hallushift:models}
\end{subtable}

    \caption{HalluShift++ performance.}
    \label{tab:hallushift}
\end{table}

As \Cref{tab:hallushift} shows, usually, we obtain low F1 scores, driven primarily by a low precision, highlighting the challenging nature of our collected data. Importantly, performances on the human-written data (bottom row of \Cref{tab:hallushift:global}) are comparable to what we observe from specific generating LVLMs. 

\begin{table}[t]
    \centering
    \resizebox{\linewidth}{!}{
    \sisetup{
table-format=1.3, round-mode=places, round-precision=3
}
\begin{tabular}{l@{{~}}>{\bf}l *{6}{S}}
\toprule
     & & \multicolumn{6}{c}{\bf calibrate on} \\
     & & {\textbf{human}} & {\textbf{gemma3}} & {\textbf{internvl}}
    & {\textbf{llava}} & {\textbf{minicpm}} & {\textbf{qwenvl}} \\
\midrule
\multirow{6}{*}{\rotatebox{90}{\bf test on}}
& human & 0.063947 & 0.056315 & 0.053490 & -0.011961 & 0.050483 & 0.069692 \\
& gemma3 & 0.070069 & 0.068898 & 0.064792 & -0.009049 & 0.061938 & 0.076707 \\
& internvl & 0.043126 & 0.040366 & 0.039185 & -0.007594 & 0.037712 & 0.044483 \\
& llava & 0.046131 & 0.047410 & 0.047458 & -0.006649 & 0.046564 & 0.041245 \\
& minicpm & 0.048230 & 0.046769 & 0.046050 & -0.007259 & 0.045216 & 0.047075 \\
& qwenvl & 0.062288 & 0.061221 & 0.060361 & -0.007612 & 0.058996 & 0.062697 \\
\bottomrule
\end{tabular}


    }
    \caption{F1 delta after recalibration (all data).}
    \label{tab:recalibration}
\end{table}
One of the reasons for the modest performances we observe in \Cref{tab:hallushift} is the high prevalence of non-hallucinated tokens. 
This prompts us to recalibrate model probabilities for non-hallucination labels by computing the optimal probability threshold in a one-versus-all setting, comparing F to all other labels. We calibrate on data derived from all possible origins (human data or any of the 5 LVLMs) and record the delta in F1 in \Cref{tab:recalibration}.
This usually leads to improvements, with one exception: calibration on Llava systematically degrades performances. As this LVLM is also the backbone for the detection pipeline, this appears to illustrate the biases that model-dependent benchmarks can introduce.

\section{Discussion}
The core question we address in this work is whether human-written samples can be used to evaluate hallucination detection models. 
To this end, we have constructed a large dataset of vision-language hallucination samples using three sampling strategies: random sampling, model-assisted pre-selection, and human-written items. 
Our analyses provide evidence that human-written samples are a viable basis for evaluating hallucination detection. 
First, they allow us to more carefully control the actual contents of our dataset
(\Cref{fig:external_labels}). Second, annotators reach a higher degree of consensus on these items (\Cref{fig:agreement-plot}). Third, human-written samples exhibit a distinct but comparable distributional profile relative to LVLM responses (\Cref{tab:f1-combined}), and the performance of hallucination detectors on human-written samples is broadly consistent with their performance on LVLM-generated data, while avoiding pitfalls that model-dependent benchmarks suffer from (\Cref{tab:recalibration}). 
It is worth stressing that the second of these points is not trivial: prior work has frequently highlighted that annotators express genuine disagreement as to what counts as a hallucination \citep{mickus-etal-2024-semeval} and where hallucination spans begin and end \citep{vazquez-etal-2025-semeval,schmidtova-etal-2026-hotelcheckspan}. While in general, some of this discrepancy can be imputed to gaps in expertise \citep{calo-etal-2026-logic}, our experimental setup allows us to cleanly isolate the effects of data sampling since the same cohort of annotators and the same inputs were used throughout our dataset.

Our work is not without its caveats.
First, we observe that the data we collect is remarkably challenging, with some models performing at or close to random on specific subsets.
Second, we cover a small number of hallucination detection models; establishing whether performance on human-written samples reliably predicts performance on true hallucinations requires broader empirical validation. Results in \Cref{tab:rankings-pearson} are encouraging, but stronger guarantees will require concerted research efforts. A shared task is currently underway to further validate this point.\footnote{ See
\href{https://helsinki-nlp.github.io/shroom/2026/}{\tt helsinki-nlp.github.io/shroom/2026/}
}
A similar comment holds for the representativeness of data derived from different LVLMs, as we do not systematically quantify variation across all possible generators.
At the same time, human-written samples can always serve as a sanity check for hallucination detection systems: because they are constructed to contain the target phenomena, a large performance gap between human-written and LVLM-generated data would likely indicate reliance on spurious signals.
Our approach alleviates some of the issues that plague modern NLP evaluation, yet it also frames hallucination detection as a black-box problem. Methods relying on internal model signals will require adaptation, as discussed in \Cref{adx:hallushift} for Hallushift++. This also reflects a broader trend toward proprietary black-box models, where evaluation must proceed without access to internal states.

It is worth acknowledging that there can be concerns as to the fitness of human-written data for evaluation purposes, as humans could introduce their own biases.\footnote{
    Bias is a valence-heavy and ambiguous word that can refer to a wide variety of phenomena --- ranging from judgments made on socionormative grounds, to distributional and structural differences between data sources (in this case, human-written vs machine-written). As far as the latter goes, we argue that \textit{some} biases can be good: e.g., it is genuinely beneficial for us to produce more instances of OCR and miscounting, as long as we do not introduce trivial distributional patterns.
}
Previous work has found genuine mismatches between human and system-based assessments, especially on disagreement modeling (e.g., \citealp{pavlick-kwiatkowski-2019-inherent,mickus-etal-2025-model}). However, such mismatches appear to exhibit a profile distinct from that observed in the present work: the gaps described in these studies appear to emerge between human and machine assessments, whereas what we target (and document in \Cref{tab:recalibration}) is bias specific to one particular model.
More broadly, prior work has documented socionormative and cognitive biases of annotators \citep{mieleszczenko-kowszewicz-etal-2023-capturing,gautam-srinath-2024-blind,blodgett-etal-2020-language} and how they can impact performances \citep{geva-etal-2019-modeling}. 
The biases observed in NLP systems tend to be exacerbated versions of human socionormative biases \citep{sheng-etal-2021-societal,chen-etal-2024-humans,10.1007/978-3-032-02728-3_49}, i.e., prior work suggests that human annotators present \textit{milder} cases of socionormative biases. 
If socionormative biases can serve as a spurious, exploitable feature that can compromise evaluation, then we should therefore expect human data to be preferable to data generated by NLP systems.
As far as SHEEP is concerned, our dataset construction sidesteps some of the issues pointed out in prior work. In particular, we keep track of annotators, hence future studies can follow \citeauthor{geva-etal-2019-modeling} and \citeauthor{gautam-srinath-2024-blind}'s recommendations.

\section{Conclusion}

We present SHEEP, a dataset of 20,000 items across four languages (Chinese, English, French, Italian) for hallucination detection across 5 LVLMs. We assess three strategies for obtaining samples: randomly selecting outputs of LVLMs, using an LVLM-as-an-annotator pre-selection, and writing samples manually. 
The dataset is richly annotated with 5 possible hallucination labels at the span level, with each item assessed independently by three annotators. The performance of modern hallucination detectors underscores the challenging nature of SHEEP.
We find that human-written samples exhibit several beneficial characteristics: they yield items that are more consensual among our annotators while maintaining a distributional profile similar to that observed in LVLM-generated samples.
We argue that this strategy avoids some pitfalls common in hallucination evaluation, including a tendency toward rapid obsolescence as models quickly cease to represent the state of the field.

\section*{Limitations}

We frame this study as a first outlook into whether human-written data could replace LVLM-generated data for hallucination detection benchmarking.
As such, there are clear limitations due to the exploratory nature of this project:
The tasks and setups considered here are not an exhaustive or representative subset of all applications and contexts where LVLM hallucination detection is relevant; we focus primarily on high-resource languages, although prior literature shows that hallucinations do not impact all languages equally \citep{vazquez-etal-2025-semeval,gamba-etal-2026-confabulations,datta-etal-2026-llm}.

An important consideration is that we propose a static dataset, hence it is in principle possible for leakage to occur. As of writing, our testset labels are kept private and accessible through a dedicated scoring interface at \url{https://shroom.pythonanywhere.com/} so as to mitigate this risk.

\section*{Ethical Considerations}
\paragraph{Data collection risks.}
The present work describes a data collection process, and therefore several ethical considerations apply. The data was deemed unlikely to present any potential harm to annotators, as inputs strictly target common objects with uncommon attributes (VISaGE, \citealp{frank-allaway-2025-visage}) or commonplace questions and images mismatched with one another (HaloQuest, \citealp{10.1007/978-3-031-72980-5_17}). The dataset contains no offensive or harmful topics, nor any personally identifying information. Annotators' usernames on the annotation platform are anonymized in the final distributed dataset.

\paragraph{Potential risks of the research agenda.} This work targets hallucination evaluation in LVLM technology. We believe that fostering a research ecosystem that actively combats the spread of misinformation and technologies that can produce misinformation at scale is of great importance. While part of this ecosystem has to include evaluation benchmarks such as the one we propose, we recognize that by design our dataset contains information that is false, incoherent, or misleading, and strictly emphasize that the intended use of this artifact is to promote research into hallucination detection and mitigation. 

\makeatletter\ifacl@finalcopy 
\section*{Acknowledgments}

The construction of this dataset was made possible thanks to a grant from the Finnish Society of Sciences and Letters.
This work has also received funding from the Digital Europe Programme under grant agreement No 101195233 (OpenEuroLLM). 
The contents of this publication are the sole responsibility of its authors and do not necessarily reflect the opinion of the EU.

\fi\makeatother

\bibliography{custom}

\appendix

\makeatletter\ifacl@finalcopy 
\section{Contributions of individual authors}
\noindent\textbf{Timothee Mickus:} overall organization, experimental designs, annotator recruitment, inter-annotator agreement (free-marginal for presence agreement), span annotation platform development, guideline development, data selection, data creation (English).

\noindent\textbf{Claudio Savelli:} HalluShift++ implementation, translation pipeline development, code development, guideline development, data creation (Italian).

\noindent\textbf{Eduardo Calò:} Theoretical framework, state of the art surveying, guideline development, data creation (Italian).

\noindent\textbf{Emilio Raimond:} LLM-as-a-judge implementation, code development, state of the art surveying, data creation (French).

\noindent\textbf{Stella Frank:} Guideline development, data selection, data creation (English), experimental design.

\noindent\textbf{Hengyu Luo:} Data creation (Chinese), translation pipeline development, code development,

\noindent\textbf{Flavio Giobergia:} Inter-annotator agreement, data quality checking.

\noindent\textbf{Vincent Segonne:} Data creation (French), translation pipeline development, sample-writing platform development.

\noindent\textbf{Chuyuan Li:} Distributional shift analysis.

\noindent\textbf{Aman Sinha:} EUQ implementation.

\noindent\textbf{Lorenzo Vaiani:} Code development.

\noindent\textbf{Jörg Tiedemann:} Organizational support.

\noindent\textbf{Raúl Vázquez:} overall organization, experimental designs, annotator recruitment, annotator briefing, span annotation platform development, guideline development, data selection.

\fi

\section{Reproducibility details}

\subsection{Translation pipeline}
\label{adx:translation}

\paragraph{Model selection.}
We evaluated candidate translation models by manually assessing 100 randomly sampled instances per target language (Italian, French, and Chinese) to measure adequacy and fluency.
Based on this evaluation, we adopt a hybrid strategy. \textbf{NLLB-200 (3.3B)} 
\citep{nllbteam2022languageleftbehindscaling} is used for translations between English and Romance languages (Italian and French), where it performs reliably. For Chinese, we instead use \textbf{Qwen3 8B} \citep{qwen3technicalreport}, which we found produces more fluent, semantically accurate outputs, particularly for short prompts and question-like inputs.

\begin{figure*}[t]
    \centering
    \begin{subfigure}[t]{\textwidth}
    \begin{instructionbox}
You are a translation engine. Translate only. Never explain, never reason, never add notes. Output only the final translation text. \\
\end{instructionbox}
\vspace{-0.25cm}
\caption{System prompt.}
\end{subfigure}
\vspace{0.25cm}

    \begin{subfigure}[t]{\textwidth}
    \begin{instructionbox}
You are a translation engine. Translate SOURCE\_TEXT into TARGET\_LANGUAGE. \\
Rules:

    1) Translate exactly what is written.
    
    2) If SOURCE\_TEXT is a question, translate the question; do NOT answer it.
    
    3) Keep named entities, numbers, URLs, and facts unchanged unless translation requires adaptation.
    
    4) Output only the translated text.
    
    5) Do not add explanations, notes, or extra lines.

SOURCE\_TEXT: \texttt{<input text>}
    \end{instructionbox}
    \vspace{-0.25cm}
    \caption{User prompt template.}
\end{subfigure}
    \caption{Prompt template used for Qwen3 translation. \texttt{TARGET\_LANGUAGE} is replaced with the desired language (e.g., Italian, French, Chinese), and \texttt{<input text>} with the source string.}
    \label{fig:translation-prompt}
\end{figure*}

\paragraph{Translation settings.}
We apply the translation pipeline in two scenarios:
(i) \textbf{English prompt translation}, where all prompts and references are translated from English into the other three languages, as both datasets are originally in English solely.
(ii) \textbf{multilingual augmentation}, used for human-written samples, where inputs may originate in any of the four languages and are translated into the remaining three. All these samples are manually checked and post-edited when necessary, ensuring high-quality multilingual consistency.

In the second setting, Qwen3 is used for all translation directions involving Chinese ($\leftrightarrow$), while NLLB is used for directions that do not involve Chinese (e.g., Italian $\leftrightarrow$ French or English).

\paragraph{NLLB pipeline.}
For Italian and French (and more generally for translations not involving Chinese), we use a sequence-to-sequence pipeline based on NLLB-200. We explicitly specify source and target language codes (e.g., \texttt{eng\_Latn} $\rightarrow$ \texttt{ita\_Latn}) and enforce the target language via a forced beginning-of-sequence token. Decoding is performed using beam search with 4 beams and no sampling.

\paragraph{Qwen pipeline.}
For Chinese and all translation directions involving Chinese, we use Qwen3 8B in an instruction-following setup. We design a structured prompt that enforces strict translation behavior (i.e., no explanations, no additional text, and no question answering). The full prompt template is reported in \Cref{fig:translation-prompt}.
We use the tokenizer chat template and perform decoding with beam search (4 beams) and no sampling to ensure deterministic outputs.

\subsection{Gemma3 model-assisted pre-selection prompt}
We provide the exact prompts used in our experiments in \Cref{fig:prompt-llm-judge}.
Because the purpose of the judge is to support large-scale pre-selection rather than to provide final annotations, we ask it to output only the predicted label, without a reasoning trace.

\subsection{Guidelines \& Annotation interfaces}
\label{adx:guidelines}

\paragraph{Guidelines for annotators and writers.}
In this appendix, we include a copy of
the guidelines provided to human writers (\Cref{fig:guidelines-human-written}) and annotators (\Cref{fig:guidelines-annot-part1,fig:guidelines-annot-part2}). The annotators had access to their guideline as a single document (split across two pages in the present reproductions for pagination purposes); minor details such as a table of contents have been omitted. Annotators also had access to a user guide for the annotation platform they were required to use, which is not reproduced here.

\begin{figure*}
    \centering
    \begin{instructionbox}
Your role is to assign a label to the provided text and image based on whether or not the text contains a *hallucination*. \\
Hallucinations correspond to cases where the provided text states information that contradicts the image. \\

LABELING INSTRUCTIONS: \\
You must assign one of the following labels (A, B, C, D, E or F):  \\

A. **Invention.** The text refers to an entity, object, property, or event that is not present in the image at all. This includes but is not limited to: fabricated objects, invented people, imagined actions unsupported by the image \\
Examples: \\
- "red bus" (no bus visible in the image) \\
- "a dog nearby" (no dog visible in the image) \\
- "an umbrella" (not present in the image)  \\

B. **Mischaracterization.** The text refers to something that is present in the image but describes it incorrectly. This includes but is not limited to: wrong object type (e.g., a taxi described as a bus), wrong colors  (e.g., a blue shirt described as green), an incorrect identity (e.g., a person standing described as sitting). \\
Examples: \\
- Calling a van a "bus" \\
- Describing a blue shirt as "green" \\
- Describing a dog running when the animal is standing or laying down.  \\

C. **Wrong Reading.** The hallucination arises from misreading text that is visible in the image. This applies whenever there is something readable in the image itself, but the provided text misquotes it, alters it, or misunderstands it. \\
Examples: \\
- the image includes a "STOP" sign but the provided text mentions a "SHOP" \\
- the image includes a "50\% OFF" sign but the provided text mentions "30\% OFF"  \\

D. **Miscounting.** The text expresses an incorrect quantity of visible items. This includes but is not limited to: wrong counts of objects shown in the image (e.g., people, cars, animals...); incorrect quantities; explicit numeric misstatements.... \\
Examples: \\
- the text refers to "three elephants" when only two elephants are visible in the image. \\
- the text refers to "many people" when only two persons are visible in the image  \\

E. **Other.** A hallucination that does not fit categories A, B, C or D.  \\

F. **No hallucination.** The provided text does not contain any hallucination: the text does not contradict the image and is perfectly coherent with the information shown in the image.  \\

Do not explain your reasoning. Only reply with the letter corresponding to the label you assign, i.e., your answer must contain only the letter A, B, C, D, E or F.  \\

PROVIDED TEXT:
    \end{instructionbox}
    \caption{Prompt used for the LLM-judge (Gemma3 27B IT).}
    \label{fig:prompt-llm-judge}
\end{figure*}

\begin{figure*}
    \centering
    \begin{instructionbox}
This document outlines guidelines for creating samples for the human-written section of DATASET. \\

\textbf{Motivation} \\
One of the things we want to verify with DATASET is whether we can evaluate the quality of hallucination detectors independently from any specific LLM output. To that end we intend to include a human-written subset of outputs in the test set, so as to be able to compare performances on LLM-generated and human-written hallucinations. \\

\textbf{Overview of the task} \\
You will be writing answers to a series of provided questions grounded on images. You will need to imitate the style of LLMs and make sure that the answer you provide contains a hallucination. Read these guidelines carefully and in their entirety before you start \\

\textbf{You are not allowed to use an LLM to generate your answers.} Answers should be written from scratch. \\

All other rules are provided as guidance: use your best judgment, and leave comments on this document whenever relevant. \\

\textbf{Detailed instructions} 
\begin{itemize}[nosep]
    \item Before you start, read the guidelines for annotators to keep in mind how the data will be presented to annotators and what task they will need to perform: \\
    <URL> \\
this document also contains the following items, which you will need to be familiar with:
\begin{itemize}[nosep]
    \item a definition of what hallucinations are
    \item a taxonomy of possible types of hallucinations (labels A, B, C, D or E)
\end{itemize}
\item Written samples are to be collected through the dedicated annotation platform, available from our private github repo: <URL>
\item Your written samples should be plausible answers to the questions as provided. Do not modify the questions.
\item Each answer you write should contain a mistake corresponding to one of the four specific labels A, B, C or D (\emph{but not E}). For instance, if your answer contains a mistake that pertains to an OCR problem (label C), it should not also include a made up entity (label A)
\item Your answer must otherwise be coherent, grammatical, and self-consistent.
\item Across all answers that you produce, you should target a balanced proportion of all four labels (25\% of A, 25\% of B, 25\% of C, and 25\% of D)
Your mistakes must be discoverable given the image. The annotator should not have to perform an external google search to spot the mistake you included.
\end{itemize} 
\null

In terms of style:
\begin{itemize}[nosep]
    \item Before you start writing samples, familiarize yourself with LLM style and the type of hallucinations they make. You can use the self-hosted pilot interface available at: <URL>
    \item Do not look at samples written by other organizers, as much as possible.
    \item Produce diverse samples. 
    \begin{itemize}[nosep]
        \item The mistakes you include do not have to target the most salient part of the question or image. For instance, consider including mistakes that target
        \begin{itemize}[nosep]
            \item objects in the background,
            \item spatial relationships between objects
            \item superfluous claims not directly/immediately relevant to the question prompt
        \end{itemize}
        \item Hallucinations can often be triggered by surprising images that depict non-standard or otherwise exceptional situations (such as three-legged cats, red bananas, cows wearing Christmas trees on their heads…), in which case LLMs can either fabricate completely ungrounded information or generate text corresponding to the most probable situation (e.g., “this cat has four legs”). Consider whether the image provided could confuse an LLM.
        \item Consider imitating different styles of LLMs, such as
        \begin{itemize}[nosep]
            \item samples with or without markdown
            \item samples with or without a sycophantic opening line (“excellent question!”)
            \item samples with or without explicit markers of reasoning (“To answer this question, I will carefully analyze all components of the image”).
        \end{itemize}
        \item Your answers can be of varying length, from a single sentence to a few paragraphs. They should not be overly short, since this would indicate that there is not much in the answer aside from the hallucination.
    \end{itemize}
    \end{itemize}

    \end{instructionbox}
    \caption{Guidelines for human-written samples production. Actual URLs are replaced with the placeholder \texttt{<URL>}.}
    \label{fig:guidelines-human-written}
\end{figure*}

\begin{figure*}
    \centering
    \begin{instructionbox}
{{\larger \textbf{Introduction}}}\\
In this annotation project you will be shown a series of question-answer pairs, plus a relevant image. The answer will be a passage of text produced by a Large Language Model (LLM) in response to the question. You will be asked to identify, with respect to the image, which spans of text in the answer constitute a “hallucination”, i.e., which parts of the answer are not supported by the image, given the question.\\
Each datapoint consists of:
\begin{itemize}[nosep]
    \item An image
    \item A question related to the image (i.e., a prompt)
    \item A response produced by a model
\end{itemize}
\null

{\larger \textbf{General Instructions}}
\begin{enumerate}[nosep]
    \item     Carefully examine the image, and read the prompt, and model response.
    \item Highlight each span of text in the model response that is not supported by the information in the image (i.e., contains a hallucination or overgeneration). Base this exclusively on the definition of hallucination we are using (see below). \\
    Important: 
    \begin{itemize}[nosep]
        \item     Annotate conservatively: annotations must include only the minimum number of characters that would need to be edited or deleted in order to make the answer correct.
        \item Prefer highlighting content words over function words.
        \item Avoid selecting full sentences unless absolutely necessary.
    \end{itemize}
    These are not rigid rules, i.e., use your best judgment when needed.
    \item Classify each marked span using the proposed taxonomy (see below). Each span must be marked with only one class label (A, B, C, D, or E).
    \item If a response contains multiple hallucinations:
    \begin{itemize}[nosep]
        \item  Highlight each span separately.
        \item Assign one label (A–E) per span. Do not merge unrelated errors into a single span.
        \item Do not overlap or embed spans.
    \end{itemize}
    \item     If a response contains no hallucination:
    \begin{itemize}[nosep]
        \item write “No hallucination” (in English) in the comment box.
    \end{itemize}
\end{enumerate}
You can also use the comment box to mark anything else that you want to raise to our attention, such as:
\begin{itemize}[nosep]
    \item model answers that are ungrammatical,
    \item outputs that are generated in the wrong language,
    \item cases where you are especially uncertain,
    \item anything else you deem relevant.
\end{itemize}
Please write comments in English, so that the whole team can read them.\\

{\larger \textbf{What counts as a hallucination?}}\\
Definition. Content that contains or describes facts that are not supported by the image. In other words: hallucinations are cases where the answer text is inaccurate, or more specific than it should be, given the provided image.\\
A hallucination is any content that:
\begin{itemize}[nosep]
    \item Refers to something not present in the image
    \item Misdescribes something visible in the image
    \item Incorrectly counts visible entities
    \item Misreads visible text
    \item Or otherwise goes beyond what can be visually grounded
\end{itemize}
\null

\textbf{What to do with unverifiable statements \& expressions of uncertainty in model outputs?} \\
Note that some model responses introduce information that cannot be verified from the image alone. These are statements that add external facts, background explanations, or narrative details that cannot be inferred directly from the image. \\
If the statement introduces new factual information that cannot be verified from the image, it should generally be treated as a hallucination. For example:
\begin{itemize}[nosep]
    \item Hallucination to annotate: statements of factual information not visible in the image (“This is a beach in Thailand” for a generic-looking tropical beach).
    \item Plausible visual inference: do not annotate: “This is a street in Paris” when the Eiffel Tower is in the background.
\end{itemize}
Model responses may also include uncertainty-marking language (e.g., “it looks like”, “it seems”, “possibly”, “it might be”). In such cases, do not annotate the uncertainty phrase itself. Instead, evaluate whether the underlying claim is visually supported. \\
Examples:
\begin{itemize}[nosep]
    \item "The man is a teacher." -> highlight "teacher" (unless the image very clearly indicates this is the case, for example if he is standing in front of a classroom of students.)
    \item "This restaurant serves Mexican food." -> highlight "Mexican" (when the restaurant in the image could serve other types of food, there is no recognizable food in sight nor any other sign of the restaurant being Mexican)
    \item "He looks sad because he lost his job" in an image with clear visual cues connecting sadness to job loss. -> DO NOT HIGHLIGHT (Not a hallucination: need to consider "sadness" being supported by facial expression and posture, Job loss is visually grounded by: a termination notice, packing office belongings into a box, etc.)
\end{itemize}

    \end{instructionbox}
    \caption{Annotator instructions, part 1 of 2: overview and general definitions.}
    \label{fig:guidelines-annot-part1}
\end{figure*}

\begin{figure*}
    \centering
\begin{instructionbox}
{{\larger \textbf{Proposed taxonomy of hallucinations}}} \\
This classification consists of 5 classes of hallucinations. \\

\textbf{A. Invention} \\
The highlighted span refers to an entity, object, property, or event that is not present in the image at all. \\
Includes: fabricated objects, invented people, imagined actions unsupported by the image \\
Examples:
\begin{itemize}[nosep]
    \item "red bus" (no bus visible)
    \item "a dog nearby" (no dog visible)
    \item "an umbrella" (not present)
\end{itemize}
Decision rule: If the entity does not exist in the image choose A \\

\textbf{B. Identity incongruity (Mischaracterization)} \\
The span refers to something that is present in the image but describes it incorrectly. \\
Includes:
\begin{itemize}[nosep]
    \item Wrong object type (taxi -> bus)
    \item Wrong color (blue shirt -> green)
    \item Wrong identity (man -> woman)
    \item Wrong activity (standing -> sitting)
    \item Incorrect spatial relations between objects or entities (below -> above, right -> left)
\end{itemize}
Examples:
\begin{itemize}[nosep]
    \item Calling a van a "bus"
    \item Describing a blue shirt as "green"
    \item Describing a dog running when the animal is standing or laying down
    \item Saying a cup is on the table when it is actually in someone’s hand
    \item Saying a person is behind the car when they are standing next to it
\end{itemize}
Decision rule: If the object exists but is misidentified or misdescribed choose B \\

\textbf{C. Wrong Reading / OCR Problem} \\
The hallucination arises from misreading text that is actually visible in the image. \\
Applies only if there is readable text in the image and the model misreads or alters it. \\
Examples:
\begin{itemize}[nosep]
    \item Image: "STOP" but response has to do with "SHOP"
    \item Image: "50\% OFF" but response is about "30\% OFF"
\end{itemize}
Do NOT use C in case:
\begin{itemize}[nosep]
    \item The text does not exist in the image. Use A in this case
    \item The error is numerical counting (not text reading). Use D in this case
\end{itemize}
\null

\textbf{D. Numeric Discrepancy / Miscounting} \\
The span expresses an incorrect quantity of visible items. \\
Includes:
\begin{itemize}[nosep]
    \item Wrong count of people, cars, animals
    \item Incorrect quantities
    \item Explicit numeric misstatements
\end{itemize}
Examples:
\begin{itemize}[nosep]
    \item "three people" (only two visible)
    \item "two dogs" (only one visible)
    \item "N something" (the thing is uncountable)
\end{itemize}
Do NOT use D in case: If a number is misread from visible written text. Use C instead. \\

\textbf{E. Other (Use Sparingly)} \\
A hallucination that does not fit categories A to D. Try to use as minimally as possible. \\
When using this tag, please leave a comment indicating why you have used it.

\end{instructionbox}
    \caption{Annotator instructions, part 2 of 2: classification of hallucinations.}
    \label{fig:guidelines-annot-part2}
\end{figure*}

\paragraph{Annotation interface}
\label{adx:interface}

\begin{figure*}
    \centering
    \includegraphics[width=0.85\linewidth]{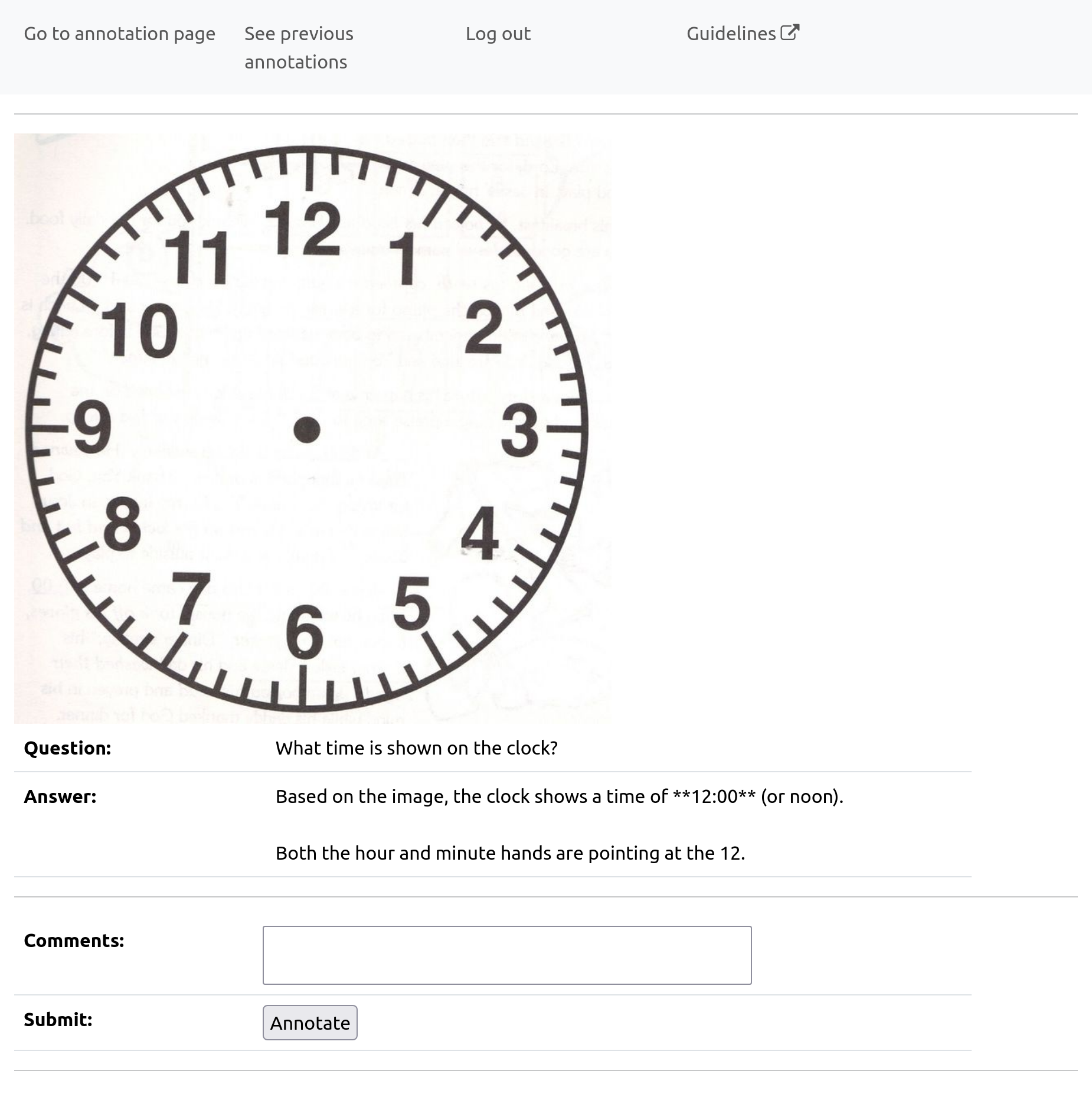}
    \caption{Screenshot of the annotation interface used (experimental item).}
    \label{fig:ui}
\end{figure*}

\begin{figure*}
    \centering
    \includegraphics[width=0.85\linewidth]{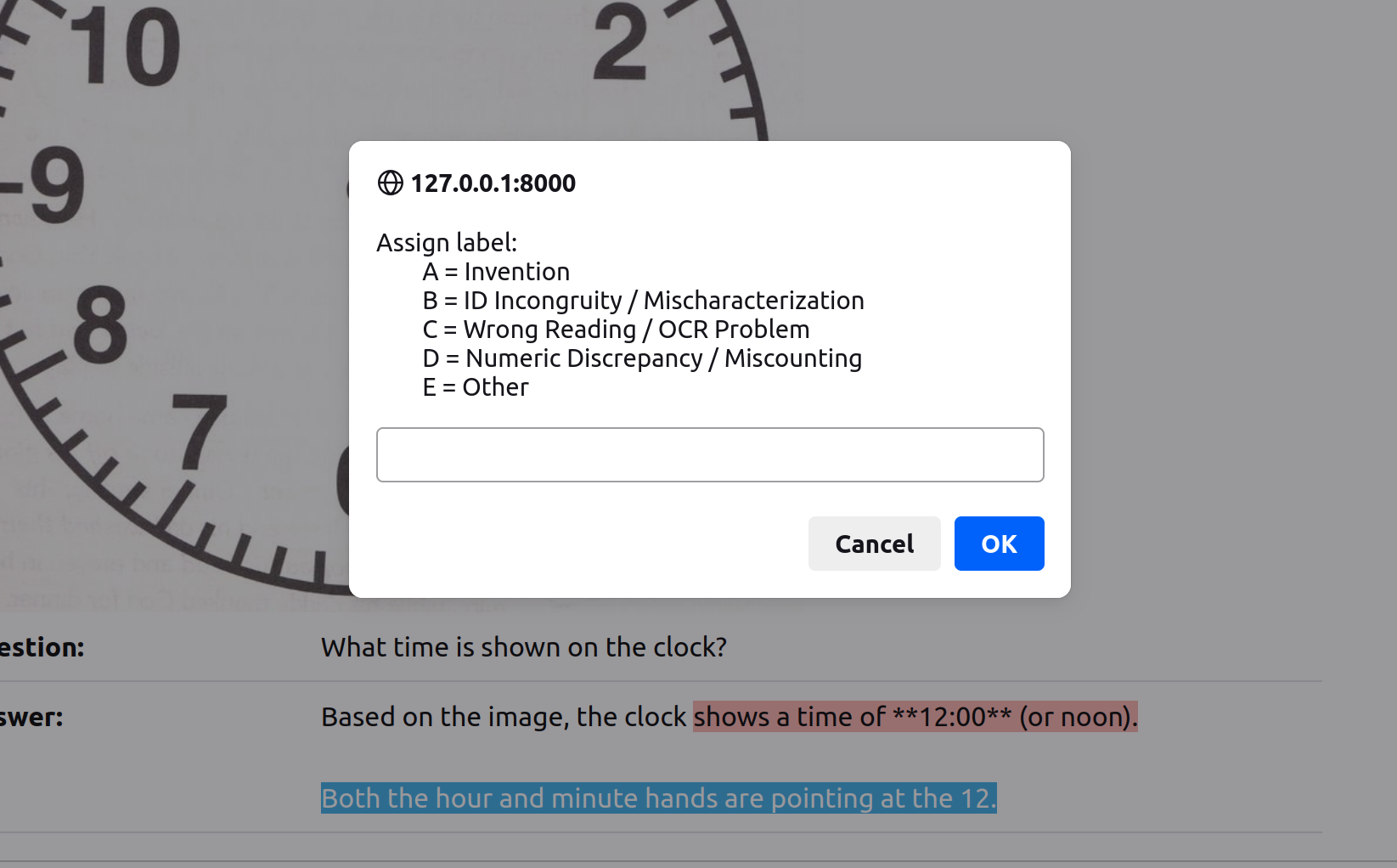}
    \caption{Screenshot of the annotation interface used (hallucination category selection box).}
    \label{fig:ui_labels}
\end{figure*}

\begin{figure*}
    \centering
    \includegraphics[width=0.85\linewidth]{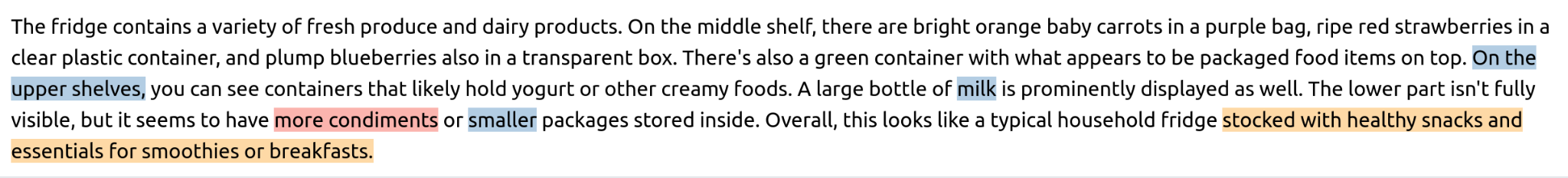}
    \caption{Screenshot of the annotation interface used (hallucination span color-coding).}
    \label{fig:ui_spans}
\end{figure*}

\begin{figure*}
    \centering
    \includegraphics[width=0.75\linewidth]{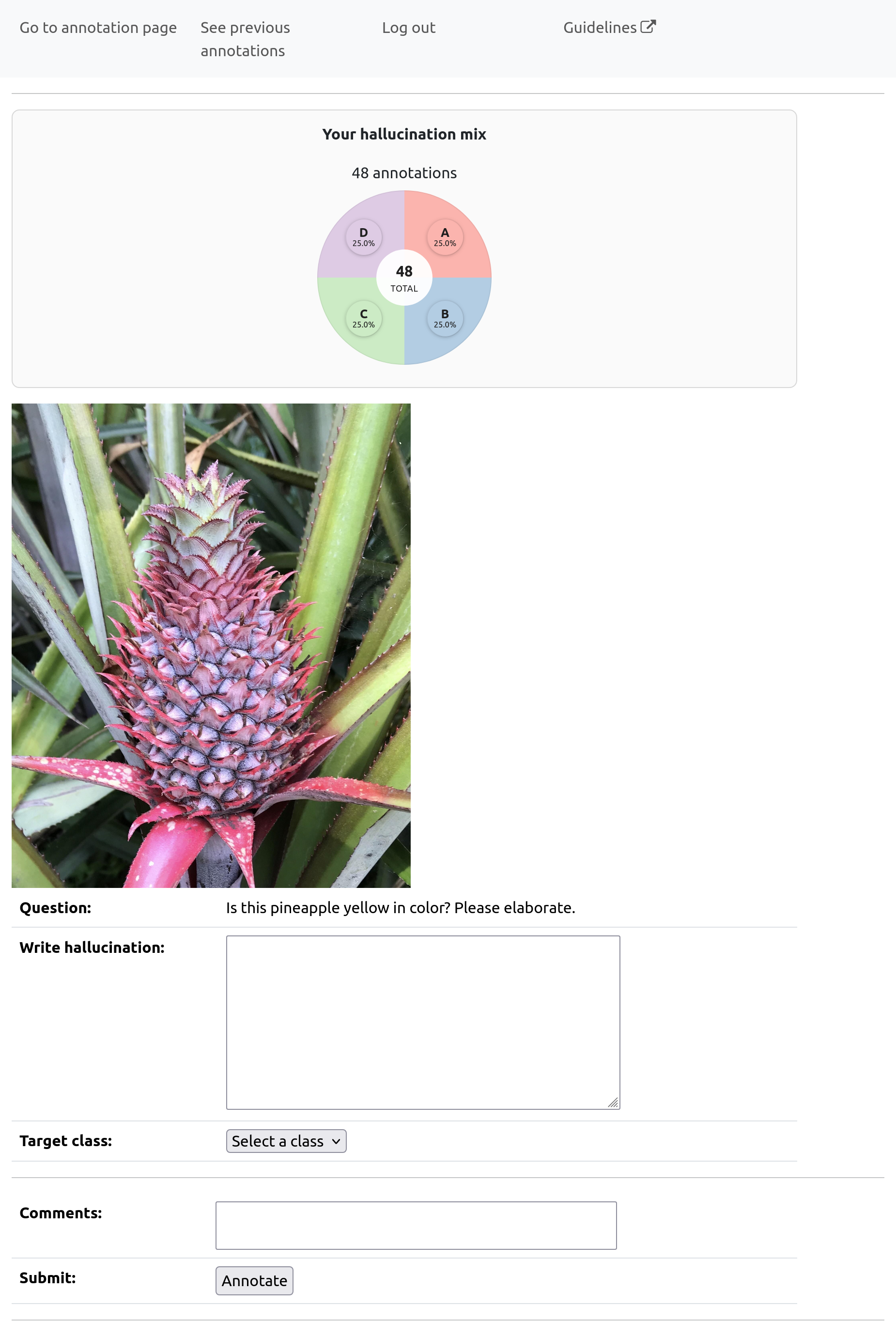}
    \caption{Screenshot of the collection interface used for human-written samples. The pie chart on top of the page provides the distribution of intended labels.}
    \label{fig:ui_human-written}
\end{figure*}

We developed a bespoke annotation interface using a stack composed of \href{https://www.djangoproject.com/}{Django 5},  \href{https://jquery.com/}{jQuery} and \href{https://getbootstrap.com/}{Bootstrap}, deployed with \href{https://nginx.org/}{NGINX} and \href{https://gunicorn.org/}{Gunicorn}. The platform was hosted on a server and used by annotators both for the training phase and for the actual annotation experiment. \Cref{fig:ui} provides a screenshot of the annotation interface for an experimental item, including the image, question, and answer. Annotators could highlight spans directly in the answer text. Upon highlighting a span, a box appeared asking them to assign one label to the selected span (\Cref{fig:ui_labels}). Annotated spans were color-coded according to their assigned labels (\Cref{fig:ui_spans}). Annotators were also free to leave additional remarks in a dedicated comment box.
A similar platform with minor modifications was used to collect human-written samples, as shown in \Cref{fig:ui_human-written}.

\section{Supplementary results}
\label{adx:sup res}

\begin{figure}[!t]
    \centering
    \includegraphics[width=\linewidth]{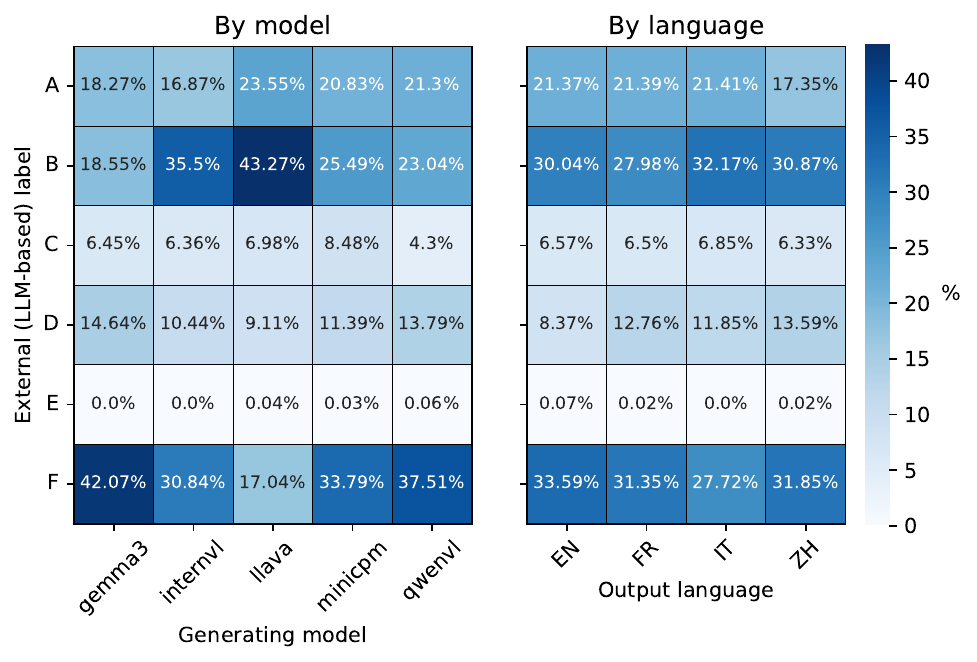}
    \caption{Distribution of labels assigned by an LLM judge (Gemma3).}
    \label{fig:prelabels}
\end{figure}

\begin{table*}[t]
\centering
\small

\begin{tabular}{l cc cc cc}
\toprule
 & \multicolumn{2}{c}{\textbf{Label agreement (longest)}}
 & \multicolumn{2}{c}{\textbf{Label agreement (first)}}
  \\
\cmidrule(lr){2-3} \cmidrule(lr){4-5} 
 & \textbf{All labels} & \textbf{Hallu. only}
 & \textbf{All labels} & \textbf{Hallu. only} \\
\midrule
\textit{\textbf{Random}}  & $0.3346$ ± $0.0067$ & $0.3453$ ± $0.0088$ & $0.3716$ ± $0.0150$ & $0.4328$ ± $0.0148$  \\
\textit{\textbf{MAP}}  & $0.3925$ ± $0.0102$ & $0.4126$ ± $0.0089$ & $0.4475$ ± $0.0141$ & $0.5020$ ± $0.0113$  \\
\textit{\textbf{Human}}  & $0.4395$ ± $0.0149$ & $0.4713$ ± $0.0163$ & $0.5881$ ± $0.0145$ & $0.6318$ ± $0.0189$  \\

\midrule
\textbf{model=gemma3}  & $0.4427$ ± $0.0142$ & $0.4611$ ± $0.0134$ & $0.5089$ ± $0.0198$ & $0.5385$ ± $0.0164$  \\
\textbf{model=internvl}  & $0.4114$ ± $0.0143$ & $0.4282$ ± $0.0142$ & $0.4595$ ± $0.0226$ & $0.4866$ ± $0.0222$  \\
\textbf{model=llava}  & $0.3419$ ± $0.0122$ & $0.3730$ ± $0.0127$ & $0.3810$ ± $0.0149$ & $0.4463$ ± $0.0188$  \\
\textbf{model=minicpm}  & $0.3914$ ± $0.0113$ & $0.4049$ ± $0.0116$ & $0.4036$ ± $0.0201$ & $0.4746$ ± $0.0253$  \\
\textbf{model=qwenvl}  & $0.2626$ ± $0.0109$ & $0.2786$ ± $0.0086$ & $0.3628$ ± $0.0205$ & $0.4289$ ± $0.0183$ \\

\midrule

\textbf{language=EN}  & $0.3698$ ± $0.0103$ & $0.4033$ ± $0.0068$ & $0.3871$ ± $0.0111$ & $0.4501$ ± $0.0143$  \\
\textbf{language=FR}  & $0.3104$ ± $0.0091$ & $0.3204$ ± $0.0063$ & $0.4119$ ± $0.0252$ & $0.4592$ ± $0.0273$  \\
\textbf{language=IT}  & $0.3963$ ± $0.0126$ & $0.4090$ ± $0.0146$ & $0.4123$ ± $0.0205$ & $0.4599$ ± $0.0194$  \\
\textbf{language=ZH}  & $0.4221$ ± $0.0092$ & $0.4380$ ± $0.0116$ & $0.4785$ ± $0.0240$ & $0.5360$ ± $0.0170$  \\

\bottomrule

\end{tabular}

\caption{Effects of agglomeration policy for multi-span annotations on label annotator agreement for \textit{Model-assisted pre-selected (MAP)}, \textit{random} and \textit{human} data, as well as separately for generating model, and output language (on MAP + random data). }
\label{tab:label-agreement-first-longest}
\end{table*}

\subsection{Detailed results for IAA assessments}
\label{adx:sup res:full IAA}

In \Cref{tab:agreement-strategies}, we present IAA measurements per strategy; in \Cref{tab:agreement-models-languages}, per language and model. These correspond to \Cref{fig:agreement-plot} (left) and  \Cref{fig:agreement-plot} (center, right) respectively.

\paragraph{Effect of label aggregation process on IAA.}
\label{adx:sup res:first-vs-longest}

In \Cref{sec:results:qc}, we mention that, when computing label agreement, we aggregate observations from annotators by simply selecting the label of the first span provided by an annotator.
An alternative strategy consists of selecting the label of the longest span instead, in terms of character coverage. A comparison of these two strategies is provided in \Cref{tab:label-agreement-first-longest}: as is apparent, the first-span approach yields higher IAA scores than the longest-span strategy; differences are statistically significant under our bootstrapping. It is not entirely clear to us why this is the case, we conjecture that it has to do with hallucination `snowballing' \citep{zhang-etal-2025-sirens}, which  annotators might react to in different ways.

\begin{table}[t]
\centering
\resizebox{0.9\linewidth}{!}{
\setlength{\tabcolsep}{3pt}
\sisetup{
  table-format=1.2,
  round-mode=places,
  round-precision=2
}

\begin{tabular}{
l
l
S@{\,{\(\pm\)}\,}S
S@{\,{\(\pm\)}\,}S
S@{\,{\(\pm\)}\,}S
}
\toprule
\multicolumn{2}{l}{\textbf{Agreement}} &
\multicolumn{2}{c}{\multirow{2}{*}{\textbf{Random}}} &
\multicolumn{2}{c}{\multirow{2}{*}{\textbf{MAP}}} &
\multicolumn{2}{c}{\multirow{2}{*}{\textbf{Human}}} \\
\textbf{Type} & \textbf{Metric} \\
\midrule

\multicolumn{2}{l}{\textbf{Pres.}}
& 0.4095 & 0.0060
& 0.4616 & 0.0106
& 0.5668 & 0.0147 \\[0.2cm]


\multirow{2}{*}{\textbf{Label}} & \textbf{Unconditional}
& 0.3491 & 0.0047
& 0.4140 & 0.0111
& 0.4710 & 0.0177 \\


& \textbf{Given a hallu.}
& 0.4270 & 0.0132
& 0.4971 & 0.0071
& 0.6342 & 0.0251 \\[0.2cm]

\multirow{2}{*}{\textbf{Span}} & \textbf{Requiring label match}
& 0.2581 & 0.0056
& 0.3133 & 0.0043
& 0.4447 & 0.0170 \\

& \textbf{Ignoring labels}
& 0.3596 & 0.0069
& 0.4210 & 0.0060
& 0.5387 & 0.0066 \\

\bottomrule
\end{tabular}}
\caption{Annotator agreement for \textit{Random}, \textit{MAP}, and \textit{Human} data. 
Values reported as mean $\pm$ 95\% bootstrap confidence interval, sampling items with replacement.}
\label{tab:agreement-strategies}
\end{table}

\begin{table*}[t]
\centering
\small
\setlength{\tabcolsep}{2pt}
\begin{adjustbox}{max width=\textwidth}
\sisetup{
  table-format=1.2,
  round-mode=places,
  round-precision=2
}

\begin{tabular}{
l
S@{\,{\(\pm\)}\,}S
S@{\,{\(\pm\)}\,}S
S@{\,{\(\pm\)}\,}S
S@{\,{\(\pm\)}\,}S
S@{\,{\(\pm\)}\,}S
@{\qquad}
S@{\,{\(\pm\)}\,}S
S@{\,{\(\pm\)}\,}S
S@{\,{\(\pm\)}\,}S
S@{\,{\(\pm\)}\,}S
}
\toprule
&
\multicolumn{10}{c}{\textbf{Generating LVLM}} &
\multicolumn{8}{c}{\textbf{Language}} \\
\cmidrule(lr){2-11}
\cmidrule(lr){12-19}
\textbf{Metric} &
\multicolumn{2}{c}{\textbf{gemma3}} &
\multicolumn{2}{c}{\textbf{internvl}} &
\multicolumn{2}{c}{\textbf{llava}} &
\multicolumn{2}{c}{\textbf{minicpm}} &
\multicolumn{2}{c}{\textbf{qwenvl}} &
\multicolumn{2}{c}{\textbf{EN}} &
\multicolumn{2}{c}{\textbf{FR}} &
\multicolumn{2}{c}{\textbf{IT}} &
\multicolumn{2}{c}{\textbf{ZH}} \\
\midrule

\textbf{Presence}
& 0.5013 & 0.0241
& 0.4885 & 0.0170
& 0.4260 & 0.0098
& 0.4718 & 0.0121
& 0.3009 & 0.0115
& 0.4664 & 0.0165
& 0.3236 & 0.0165
& 0.4616 & 0.0066
& 0.4950 & 0.0178 \\

\textbf{Label (longest/all)}
& 0.4405 & 0.0101
& 0.4242 & 0.0147
& 0.3417 & 0.0075
& 0.3854 & 0.0124
& 0.2634 & 0.0145
& 0.3714 & 0.0093
& 0.3051 & 0.0083
& 0.3891 & 0.0093
& 0.4275 & 0.0165 \\

\textbf{Label (first/all)}
& 0.4596 & 0.0140
& 0.4262 & 0.0119
& 0.3758 & 0.0186
& 0.4126 & 0.0113
& 0.2761 & 0.0198
& 0.3985 & 0.0069
& 0.3261 & 0.0113
& 0.4081 & 0.0116
& 0.4390 & 0.0091 \\

\textbf{Label (longest/hallu.)}
& 0.4919 & 0.0261
& 0.4691 & 0.0212
& 0.3850 & 0.0183
& 0.4176 & 0.0134
& 0.3673 & 0.0123
& 0.3794 & 0.0132
& 0.4165 & 0.0108
& 0.4236 & 0.0220
& 0.4877 & 0.0157 \\

\textbf{Label (first/hallu.)}
& 0.5391 & 0.0299
& 0.4958 & 0.0201
& 0.4517 & 0.0153
& 0.4765 & 0.0119
& 0.4215 & 0.0277
& 0.4598 & 0.0182
& 0.4449 & 0.0178
& 0.4697 & 0.0139
& 0.5368 & 0.0114 \\

\textbf{Span (ignoring labels)}
& 0.4364 & 0.0186
& 0.4486 & 0.0107
& 0.3884 & 0.0086
& 0.3799 & 0.0054
& 0.3220 & 0.0103
& 0.3541 & 0.0067
& 0.3865 & 0.0134
& 0.4257 & 0.0100
& 0.4251 & 0.0110 \\

\textbf{Span (Requiring label match)}
& 0.3185 & 0.0127
& 0.3277 & 0.0147
& 0.2862 & 0.0067
& 0.2897 & 0.0108
& 0.2362 & 0.0148
& 0.2528 & 0.0066
& 0.2866 & 0.0061
& 0.3117 & 0.0134
& 0.3242 & 0.0136 \\

\bottomrule
\end{tabular}
\end{adjustbox}
\caption{Annotator agreement grouped by generating LVLM and output language on the random and MAP subsets. See \Cref{tab:agreement-strategies} for an overview of the metrics. IAA scores are reported as mean $\pm$ 95\% bootstrap confidence interval.}
\label{tab:agreement-models-languages}
\end{table*}

\subsection{Supplementary results on Gemma3-assisted labeling}
\label{adx:gemma}

\begin{table}[t]
    \centering
    \resizebox{0.825\linewidth}{!}{
    \sisetup{
  table-format=1.3,
  round-mode=places,
  round-precision=3
}
\begin{tabular}{>{\bf}l>{\bf}l*{4}{S}}    
\toprule
&&{{\textbf{acc}}} & {{\textbf{prec}}} & {{\textbf{rec}}}&{{\textbf{F1}}} \\
\midrule
\multirow{6}{*}{\rotatebox{90}{overall}}
&  human          & 0.664375  & 0.575707  & 0.508243  & 0.498913 \\
&         gemma3  & 0.542763  & 0.504982  & 0.444466  & 0.442032 \\
&       internvl  & 0.531915  & 0.460329  & 0.383315  & 0.400734 \\
&          llava  & 0.574468  & 0.464656  & 0.489872  & 0.455147 \\
&        minicpm  & 0.485437  & 0.491528  & 0.393200  & 0.409673 \\
&         qwenvl  & 0.459459  & 0.503413  & 0.412179  & 0.426878 \\
\midrule
\multirow{6}{*}{\rotatebox{90}{English}}

  & human         & 0.695000 & 0.582502 & 0.543911 & 0.510781 \\
  &        gemma3 & 0.521739 & 0.685516 & 0.396531 & 0.391259 \\
  &      internvl & 0.516129 & 0.462537 & 0.303195 & 0.328004 \\
  &         llava & 0.567901 & 0.590857 & 0.544474 & 0.515805 \\
  &       minicpm & 0.448718 & 0.457937 & 0.344344 & 0.344703 \\
  &        qwenvl & 0.402174 & 0.405242 & 0.321846 & 0.306616 \\
\midrule
\multirow{6}{*}{\rotatebox{90}{French}}
  & human         & 0.677500 & 0.576933 & 0.517647 & 0.509476 \\
  &        gemma3 & 0.539474 & 0.442298 & 0.532828 & 0.449839 \\
  &      internvl & 0.463415 & 0.324603 & 0.358961 & 0.321223 \\
  &         llava & 0.538462 & 0.441282 & 0.545690 & 0.448990 \\
  &       minicpm & 0.480519 & 0.468030 & 0.415278 & 0.420988 \\
  &        qwenvl & 0.426667 & 0.513173 & 0.481401 & 0.458223 \\
\midrule
\multirow{6}{*}{\rotatebox{90}{Italian}}
  & human         & 0.650000 & 0.582411 & 0.486722 & 0.482303 \\
  &        gemma3 & 0.539474 & 0.504650 & 0.475439 & 0.451861 \\
  &      internvl & 0.618421 & 0.407343 & 0.380435 & 0.383105 \\
  &         llava & 0.676056 & 0.554082 & 0.515702 & 0.516484 \\
  &       minicpm & 0.518519 & 0.609305 & 0.412860 & 0.437302 \\
  &        qwenvl & 0.433333 & 0.450000 & 0.523016 & 0.458120 \\
\midrule
\multirow{6}{*}{\rotatebox{90}{Chinese}}
  & human         & 0.635000 & 0.562949 & 0.495067 & 0.491000 \\
  &        gemma3 & 0.566265 & 0.520266 & 0.518065 & 0.505675 \\
  &      internvl & 0.532258 & 0.596923 & 0.457955 & 0.505234 \\
  &         llava & 0.507692 & 0.365247 & 0.449151 & 0.390369 \\
  &       minicpm & 0.493151 & 0.422113 & 0.385749 & 0.393317 \\
  &        qwenvl & 0.594203 & 0.572121 & 0.371843 & 0.424621 \\
\bottomrule
\end{tabular}
}
    \caption{Gemma3 classification full results.}
    \label{tab:gemma:full}
\end{table}

\paragraph{Full classification results.}
In \Cref{tab:gemma:full}, we provide a complete overview of classification metrics for the label assessment using the Gemma3 LLM-judge. Similar observations as mentioned in the main text hold.

\paragraph{Gemma3 assessment across languages and generating LVLM.}
For LLM-based external labels, we additionally study the distribution of labels assigned by the Gemma3 judge across models and languages, as shown in Figure \ref{fig:prelabels}.  While the distribution of LLM-based assessment across languages is stable, noteworthy patterns emerge when comparing models: Gemma3 rates its output as less likely to contain hallucinations, whereas Internvl and Llava are rated as containing more mischaracterizations.

\subsection{Supplementary results on HalluShift++}
\label{adx:sup res:hallushift}

\paragraph{HalluShift++ implementation details.}
\label{adx:hallushift}
Our HalluShift++ baseline follows the intuition of \citet{nath2025hallushift++}: hallucinated content can be detected from internal model signals, including uncertainty, representation shifts, and attention behavior. Given an image, a question, and a generated answer, we perform a teacher-forced forward pass through an evaluating VLM. This evaluation model can either be the same LVLM that produced the answer, corresponding to a model-dependent self-probing setup, or a separate LVLM used to evaluate outputs generated by other models. In all cases, the original generation process remains unchanged; the forward pass is used only post hoc to extract token-level features.

For each response token, we extract uncertainty-, representation-, and attention-based features. The uncertainty features include the maximum softmax probability, its reciprocal as a confidence proxy, and the probability assigned to the observed token under teacher forcing. From the latter, we compute token-level negative log-likelihood and perplexity. The representation features include layer-wise hidden-state norms and layer-consistency scores, computed as the normalized cosine similarity between an early and a late decoder layer, together with their complements. The attention features include layer-wise attention means and entropies, as well as attention-concentration statistics based on the mean and standard deviation of Gini coefficients over the last attention layers. We additionally include sequence-level confidence and token-pattern features. These comprise the mean and standard deviation of the inverse maximum probability, the confidence trend across the answer, the mean confidence, the fraction of low-confidence tokens, repetition ratios, and normalized lexical diversity. Since our prediction target is token-level, sequence-level features are attached to every token of the corresponding answer.

To obtain token-level supervision, we align each decoded token with its character span in the generated answer. Tokens whose character span overlaps an annotated hallucination span are labeled as hallucinated; tokens with no overlap are labeled as non-hallucinated. In the multiclass setting, hallucinated tokens are assigned the corresponding hallucination type. We then train a lightweight multilayer perceptron over the resulting token-level feature vectors, using the same classifier configuration as \citet{nath2025hallushift++}. 
Our adaptation differs from the original HalluShift++ formulation in one main respect: while HalluShift++ decomposes generated descriptions into semantic chunks, such as object, attribute, and relation units, our annotations are span-based. We therefore align internal signals directly to decoded tokens and train a token-level classifier.

\begin{table}[t]
    \centering

\begin{subtable}{\linewidth}
    \centering
    \resizebox{0.9\linewidth}{!}
    {
\sisetup{
  table-format=1.3,
  round-mode=places,
  round-precision=3
}
\begin{tabular}{>{\bf}l *{4}{S}}
\toprule
Strategy &{{\bf acc }} &{{\bf prec }} &{{\bf rec }} &{{\bf F1 }} \\
\midrule
       random  & 0.658566 & 0.179725 & 0.671580 & 0.283564 \\
 MAP  & 0.633210 & 0.267726 & 0.700256 & 0.387355 \\
 human-written  & 0.565402 & 0.262836 & 0.846148 & 0.401085 \\

\bottomrule
\end{tabular}

}
\caption{HalluShift++ performance (accuracy, recall, precision, F1) across strategies. \vspace{0.125cm}}
\label{tab:hallushift:global:bin}
\end{subtable}

\begin{subtable}{\linewidth}
    \centering
    \small
\sisetup{
  table-format=1.3,
  round-mode=places,
  round-precision=3
}
\begin{tabular}{>{\bf}l *{3}{S}}
\toprule
     &  {{\bf random}} &  {{\bf MAP}} &  {{\bf all}} \\
\midrule
  gemma3  & 0.256988 & 0.418344 & 0.324460 \\
internvl  & 0.280290 & 0.382342 & 0.335759 \\
   llava  & 0.398239 & 0.401863 & 0.400587 \\
 minicpm  & 0.276398 & 0.345322 & 0.312247 \\
  qwenvl  & 0.247787 & 0.391095 & 0.326951 \\
\bottomrule
\end{tabular}

\caption{HalluShift++ F1 per generating LVLM.}
\label{tab:hallushift:models:bin}
\end{subtable}

    \caption{HalluShift++ performance (binary label-classification setup).}
    \label{tab:hallushift:bin}
\end{table}

\paragraph{Performance in binary setting.}
In \Cref{tab:hallushift:bin}, we provide an overview of performances for the HalluShift++ method based on span detection, framing the problem as a binary token-labeling task instead (i.e., F vs. all). Similar conclusions emerge: a very low precision severely curtails the performances of these models. In \Cref{tab:hallushift:models:bin}, we see a significant downgrade in performance for randomly selected items.

\subsection{Span assessment with EUQ} 
\label{adx:euq}

Another baseline we consider is based on Evidential Uncertainty Quantification \cite[EUQ;][]{huang2026detecting}. It provides a training-free framework that explicitly decomposes token-level epistemic uncertainty into different types of misbehaviors. The framework applies basic belief assignment \cite{dempster1967upper} to the token-level pre-logits feature obtained from the output head of LVLM to compute evidence weights. These weights are then decomposed into positive and negative components, which are fused to estimate the final uncertainties that can detect conflict and ignorance. Conflict refers to error caused by the difference in the image and text modality; whereas, ignorance refers to error caused by missing information. Using the optimal threshold obtained from a calibration set for conflict and ignorance, we filter spans for containing hallucinations.
\begin{table}[t]
    \centering
    \small
    \sisetup{
  table-format=1.3,
  round-mode=places,
  round-precision=3
}
\begin{tabular}{l@{{~}}>{\bf}l *{4}{S}}
\toprule
& & {{\textbf{acc}}} & {{\textbf{prec}}} & {{\textbf{rec}}} & {{\textbf{F1}}} \\
\midrule
\multirow{3}{*}{\rotatebox{90}{\bf strategy}}
&        random & 0.794425 & 0.195630 & 0.198189 & 0.196901 \\
&  MAP & 0.733111 & 0.269008 & 0.167343 & 0.206332 \\
& human-written & 0.719356 & 0.266773 & 0.214856 & 0.238017 \\
\midrule
\multirow{5}{*}{\rotatebox{90}{\bf random}}
&         gemma3 & 0.816769 & 0.179050 & 0.230009 & 0.201355 \\
&       internvl & 0.804312 & 0.201125 & 0.219325 & 0.209831 \\
&          llava & 0.708015 & 0.330688 & 0.176732 & 0.230354 \\
&        minicpm & 0.799420 & 0.190815 & 0.213286 & 0.201426 \\
&         qwenvl & 0.809396 & 0.155739 & 0.187901 & 0.170315 \\
\midrule
\multirow{5}{*}{\rotatebox{90}{\bf MAP}}
&         gemma3 & 0.753449 & 0.322938 & 0.173363 & 0.225611 \\
&       internvl & 0.756990 & 0.311334 & 0.186693 & 0.233417 \\
&          llava & 0.685957 & 0.320406 & 0.146495 & 0.201061 \\
&        minicpm & 0.764196 & 0.274089 & 0.223853 & 0.246437 \\
&         qwenvl & 0.735367 & 0.213753 & 0.152283 & 0.177857 \\
\midrule
\multirow{5}{*}{\rotatebox{90}{\bf all}}
&         gemma3 & 0.794034 & 0.225908 & 0.199641 & 0.211964 \\
&       internvl & 0.779899 & 0.255682 & 0.198420 & 0.223441 \\
&          llava & 0.693847 & 0.324248 & 0.156713 & 0.211301 \\
&        minicpm & 0.782474 & 0.232112 & 0.219350 & 0.225550 \\
&         qwenvl & 0.773602 & 0.184717 & 0.165523 & 0.174594 \\
\bottomrule
\end{tabular}
    \caption{Span detection with EUQ, token-level classification results.}
    \label{tab:euq}
\end{table}
Results are shown in \Cref{tab:euq}. The high label imbalance between hallucinated and non-hallucinated tokens leads to poor performances out-of-the-box, lower than what we observed with HalluShift++ in \Cref{tab:hallushift:bin}.
For accuracy, recall, and precision, we can always find specific LVLM-based setups that either under- or over-perform with respect to the human-written samples subset, although the human-written data does yield the highest F1 score.

\section{Disclaimers}
\paragraph{Use of existing artifacts.}
We rely on preexisting models and datasets to construct our own dataset; the use is compatible with explicit licenses.
\paragraph{Use of AI tools.}
AI technology was used to jump-start coding. Authors take full responsibility for the contents of this paper. 
\paragraph{Hardware and compute.}
Our experiments were run on a variety of compute clusters, including V100, A100, RTX A6000, MI250x, H1000. 

LVLM answer generation for all models except Gemma3 was carried out on a node equipped with two NVIDIA RTX A6000 GPUs. Generation required approximately 4.5 seconds per sample across almost all models, whereas Qwen3-VL required approximately 9 seconds per sample on average.

LVLM answer generation for Gemma3 was carried out on a node equipped with one AMD MI250x GPU. Generation required approximately 6 seconds per sample on average.

\end{document}